\documentclass{article}

     \PassOptionsToPackage{numbers, compress}{natbib}

      \usepackage[preprint]{neurips_2025}

\usepackage[utf8]{inputenc} 
\usepackage[T1]{fontenc}    
\usepackage{hyperref}       
\usepackage{url}            
\usepackage{booktabs}       
\usepackage{amsfonts}       
\usepackage{nicefrac}       
\usepackage{microtype}      
\usepackage[table]{xcolor}
\usepackage{graphicx}
\usepackage{xspace}
\usepackage{makecell}
\usepackage{adjustbox}
\usepackage{listings}
\usepackage{fancyvrb}
\usepackage{multirow}
\usepackage{hhline}
\usepackage{amsmath}
\usepackage{enumitem}
\usepackage[most]{tcolorbox}
\definecolor{SP}{RGB}{114, 97, 171}

\newcommand{\NAME}{{MMIG-Bench}\xspace}
\usepackage{pifont}

\definecolor{nipsPink}{RGB}{219, 50, 121}

\hypersetup{
    urlbordercolor=white
}

\newcommand{\pinkurl}[1]{\href{#1}{\textcolor{nipsPink}{\texttt{#1}}}}

\title{
\NAME: Towards Comprehensive and Explainable Evaluation of Multi-Modal Image Generation Models}
\author{\bf Hang Hua\textsuperscript{1}$^{}$\thanks{Equal Contribution}
\hspace{1em} \bf Ziyun Zeng\textsuperscript{1}$^{\ast}$
\hspace{1em} \bf Yizhi Song\textsuperscript{2}$^{\ast}$
\hspace{1em} \bf Yunlong Tang \textsuperscript{1}\\
\hspace{1em} \bf Liu He \textsuperscript{2} 
\hspace{1em} \bf Daniel Aliaga \textsuperscript{2}
\hspace{1em} \bf Wei Xiong \textsuperscript{3}
\hspace{1em} \bf Jiebo Luo\textsuperscript{1}$^{}$\thanks{Corresponding Author}
\\
\textsuperscript{1} University of Rochester ~~
\textsuperscript{2} Purdue University ~~
\textsuperscript{3} NVIDIA \\
\texttt{\{hhua2, jluo\}@cs.rochester.edu, \{zzeng24, ytang37\}@ur.rochester.edu, }\\
\texttt{\{song630, he425, aliaga\}@purdue.edu, wxiong@nvidia.com}
}

\begin{document}
\maketitle
\begin{figure}[ht]
    \vspace{-30pt}
    \centering
    \includegraphics[width=0.95\linewidth]{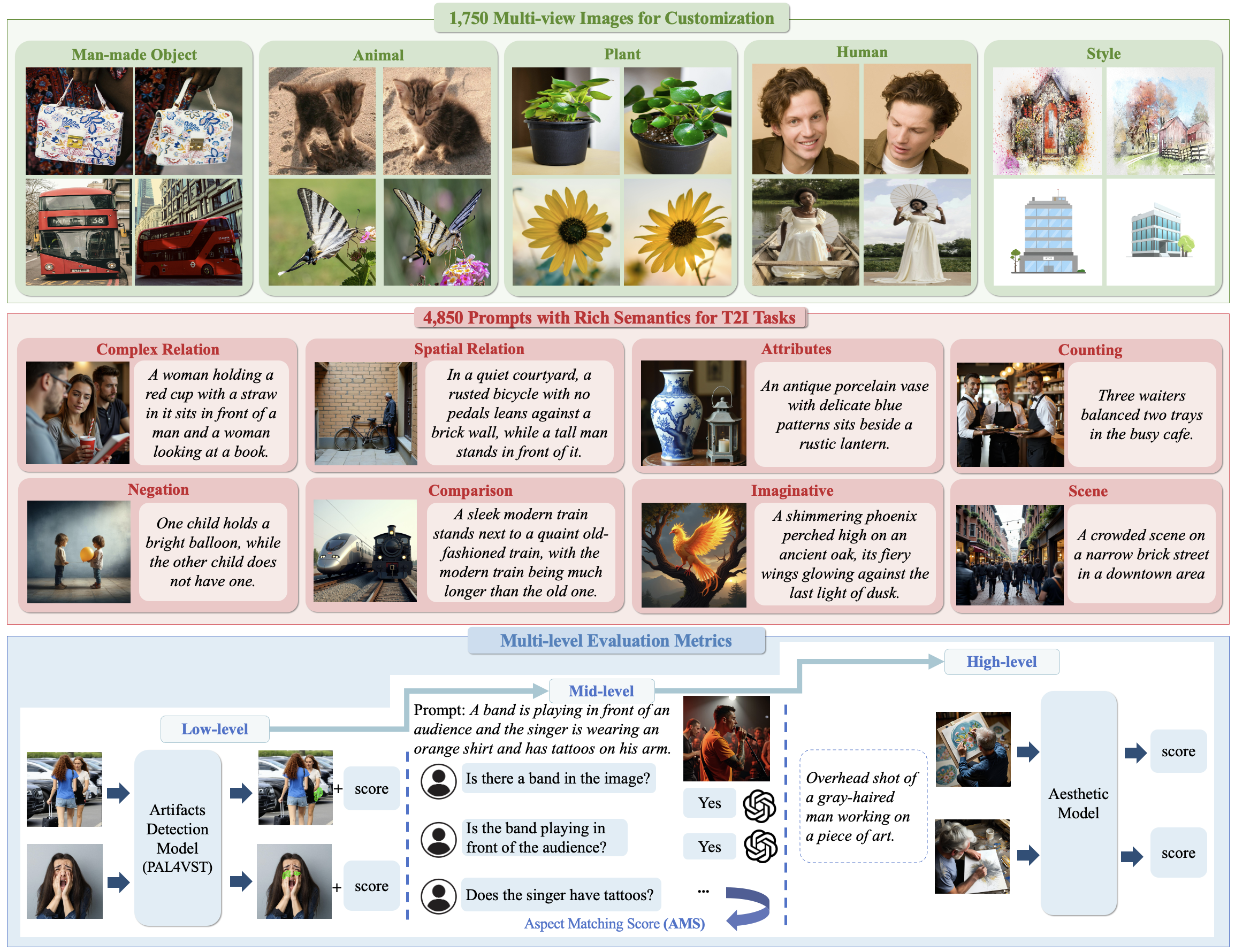}
    \vspace{-10pt}
    \caption{Overview of \NAME. We present a unified multi-modal benchmark which contains 1,750 multi-view reference images with 4,850 richly annotated text prompts, covering both text-only and image-text-conditioned generation.
    We also propose a comprehensive three-level evaluation framework: low-level of artifacts and identity preservation, mid-level of VQA-based Aspect Matching Score, and high-level of aesthetics and human preferences—delivers holistic and interpretable scores.}
    \label{fig:teaser}
\end{figure}
\definecolor{my_green}{RGB}{51,102,0}
\definecolor{my_red}{RGB}{204, 0, 0}

\begin{abstract}
Recent multimodal image generators such as GPT-4o, Gemini 2.0 Flash, and Gemini 2.5 Pro excel at following complex instructions, editing images and maintaining concept consistency.
However, they are still evaluated by \textit{disjoint} toolkits: text-to-image (T2I) benchmarks that lacks multi-modal conditioning, and customized image generation benchmarks that overlook compositional semantics and common knowledge.
We propose \NAME, a \textit{comprehensive} \textbf{M}ulti-\textbf{M}odal \textbf{I}mage \textbf{G}eneration \textbf{Bench}mark that unifies these tasks by pairing 4,850 richly annotated text prompts with 1,750 multi-view reference images across 380 subjects, spanning humans, animals, objects, and artistic styles.
\NAME is equipped with a three-level evaluation framework: (1) low-level metrics for visual artifacts and identity preservation of objects; (2) novel Aspect Matching Score (AMS): a VQA-based mid-level metric that delivers fine-grained prompt-image alignment and shows strong correlation with human judgments; and (3) high-level metrics for aesthetics and human preference.
Using \NAME, we benchmark 17 state-of-the-art models, including Gemini 2.5 Pro, FLUX, DreamBooth, and IP-Adapter, and validate our metrics with 32k human ratings, yielding in-depth insights into architecture and data design. Resources are available at: 
\pinkurl{https://hanghuacs.github.io/MMIG-Bench/}
\end{abstract}
    
\vspace{-3mm}
\section{Introduction}
\label{sec:intro}

With the rapid progress in foundational image generation systems, a diverse range of models has emerged at the forefront of research and application. These include commercial models such as GPT-4o \cite{openai2023gpt4} and Gemini 2.0 Flash, as well as opaen-source models like FLUX \cite{flux2024}, Hunyuan-DiT \cite{Li2024HunyuanDiTAP}, Emu3 \cite{Wang2024Emu3NP}, and DreamO \cite{mou2025dreamo}.
Currently, the community primarily evaluates them with separate toolkits: text-to-image (T2I) benchmarks that focus on compositionality and world knowledge; and customized generation benchmarks that emphasize identity preservation of the reference images.
However, fine-grained semantic alignment and compositional reasoning included in the T2I evaluation are equally critical for the customization task; conversely, providing reference images with text enhances the flexibility and also broadens the expressive scope of generation—enabling style transfer and other capabilities that pure T2I tasks does not contain. Therefore there is a pressing need for a comprehensive and unified benchmark that treats multi-modal input (both text and image) as an integrated entity.

To be more specific, early T2I benchmarks (e.g., PartiPrompts, Gecko) are large sparsely labelled, typically assigning only a single category per prompt. Recent benchmarks (T2I-CompBench++ \cite{Huang2023T2ICompBenchAE}, GenEval \cite{ghosh2023geneval}, GenAI-Bench \cite{Li2024GenAIBenchEA}, T2I-FactualBench \cite{huang2024t2i}) incorporate dense tags, evaluating nuanced aspects of generated images such as compositionality, common sense, and world knowledge. However, they focus on evaluating generators only conditioned on text, and thus are limited in evaluating newer multi-modal generation models with both images and text as input, such as GPT-4o and Gemini 2.0 Flash.
Customization benchmarks~\cite{dreambench, Peng2024DreamBenchAH} are still scarce, 
most are tiny and lack enough multiview reference images.
In addition, the evaluation metrics in T2I benchmarks mostly score prompt following, overlooking visual fidelity. Customization benchmarks often rely on trivial approaches to assess semantic alignment or identity preservation, lacking fine-grained and effective metrics.

To address these issues, we build the first comprehensive multi-modal benchmark \NAME for image generation. we summarize our contributions below and illustrate them in Fig.~\ref{fig:teaser} and Fig.~\ref{fig:stat}.

\begin{itemize}[left=0pt]
    \item \textbf{Unified task coverage and multi-modal input}. We collect over \textbf{380} groups (animal, object, human,, and style) comprising \textbf{1,750} multiview object-centric images enabling rigorous reference‑based generation. We also construct \textbf{4,850} richly annotated prompts across compositionality (attributes, relations, objects, and numeracy), style (fixed pattern, professional, natural, human-written), realism (imaginative) and common sense (comparisons, negations). The proposed benchmark provides future research with the flexibility to conduct any image generation task.
    \item \textbf{Three‑level evaluation suite.} We propose a multilevel scoring framework for comprehensive evaluation. (1) Low-level metrics assess visual artifacts and identity preservation of objects; (2) At mid-level, we propose the \textbf{Aspect Matching Score (AMS)} : a novel VQA-based metric that captures fine-grained semantic alignment, showing strong correlation with human perception; (3) high-level metrics measure aesthetics and human preferences. This multi‑level framework expands T2I evaluation beyond prompt adherence and provides customized generation the nuanced semantic assessment it lacks.
\end{itemize}

We validate our metrics with \textbf{32k} human ratings and benchmark 17 state-of-the-art models, offering design insights on architecture choices and data curation. We will release \NAME and the evaluation code to accelerate future research on multi-modal generation.

\section{Related Work}
\label{sec:formatting}

\subsection{Text-to-Image Generation}
\label{sec:related_work_t2i}
Recent advancements in text-to-image generation have significantly enhanced models' visual synthesis capabilities. FLUX.1-dev \cite{flux2024} employs a rectified flow transformer integrated with 3D modeling, enabling precise compositional control. Hunyuan-DiT \cite{Li2024HunyuanDiTAP} advances diffusion transformers with multilingual support and multimodal dialogue, enhancing caption accuracy. Lumina-Image 2.0 \cite{Qin2025LuminaImage2A} prioritizes efficiency through unified architectures and progressive training, achieving scalability with compact models. Photon-v1 \cite{photonv1} specifically targets photorealism, effectively rendering challenging visual elements. PixArt-$\Sigma$ \cite{Chen2024PixArtWT} innovates with attention mechanisms, achieving ultra-high-resolution generation. Stable Diffusion variants, including SDXL \cite{Podell2023SDXLIL} and SD3.5 \cite{Esser2024ScalingRF}, leverage advanced multimodal conditioning to enhance image quality and textual fidelity. Janus Pro \cite{Chen2025JanusProUM} offers superior multimodal stability through optimized training and extensive datasets. Finally, CogView4 \cite{Zheng2024CogView3FA}, with its large-scale parameters, sets benchmarks in visual fidelity and resolution, highlighting ongoing innovation in generative image synthesis.

\subsection{Customized Image Generation}
\label{sec:related_work_custom}
Customized image generation techniques have significantly advanced, enabling precise, context-specific visual content \cite{wei2025personalized}. DreamBooth \cite{ruiz2022dreambooth} and HyperDreamBooth \cite{ruiz2023hyperdreambooth} established robust frameworks for efficient subject-driven fine-tuning from minimal references. Methods like Imagic \cite{kawar2023imagic} and Textual Inversion \cite{gal2022image} embed new concepts into pretrained models for semantic editing without extensive retraining. InstantBooth \cite{shi2024instantbooth} and GroundingBooth \cite{xiong2024groundingbooth} streamline personalization, reducing computational costs and training time. Multimodal models such as Kosmos-G \cite{Pan2023KosmosGGI}, UNIMO-G \cite{li2024unimo}, and Emu3 \cite{Wang2024Emu3NP} expand personalization capabilities through multimodal integration and semantic understanding. BLIP-Diffusion \cite{li2023blip} and IP-Adapter \cite{ye2023ip} enhance visual grounding between textual prompts and personalized features. InstantID \cite{wang2024instantid} specializes in identity-aware personalization with high-fidelity identity preservation. Recently, Personalize Anything \cite{feng2025personalize} and DreamO \cite{mou2025dreamo} have further advanced the field, enabling versatile, contextually adaptive image synthesis.

\subsection{Benchmarks and Metrics for Image Generation}
\label{sec:related_work_benchmarks}
Recent benchmarks and metrics comprehensively evaluate generative image models. DreamBench++ \cite{Peng2024DreamBenchAH} and GenAI-Bench \cite{Li2024GenAIBenchEA} systematically assess generative AI across diverse tasks, while PartiPrompts \cite{Yu2022ScalingAM} and Gecko \cite{wiles2024revisiting} provide specialized datasets for prompt-based generation fidelity. T2I-CompBench and T2I-CompBench++ \cite{Huang2023T2ICompBenchAE} target compositional complexity and context understanding. DPG-Bench \cite{hu2024ella} focuses on perceptual metrics, whereas GenEval \cite{ghosh2023geneval} and HEIM \cite{lee2023holistic} offer robust frameworks for systematic comparison. Q-Eval-100K \cite{Zhang2025QEval100KEV} and T2I-FactualBench \cite{huang2024t2i} specifically evaluate factual consistency and quality alignment. Additionally, LMM4LMM \cite{Wang2025LMM4LMMBA} assesses multimodal language models for image generation, and EvalMuse-40K \cite{Han2024EvalMuse40KAR} provides extensive benchmarking of image quality and model performance.

\begin{figure}[htbp]
    \centering
    \includegraphics[width=1.0\linewidth]{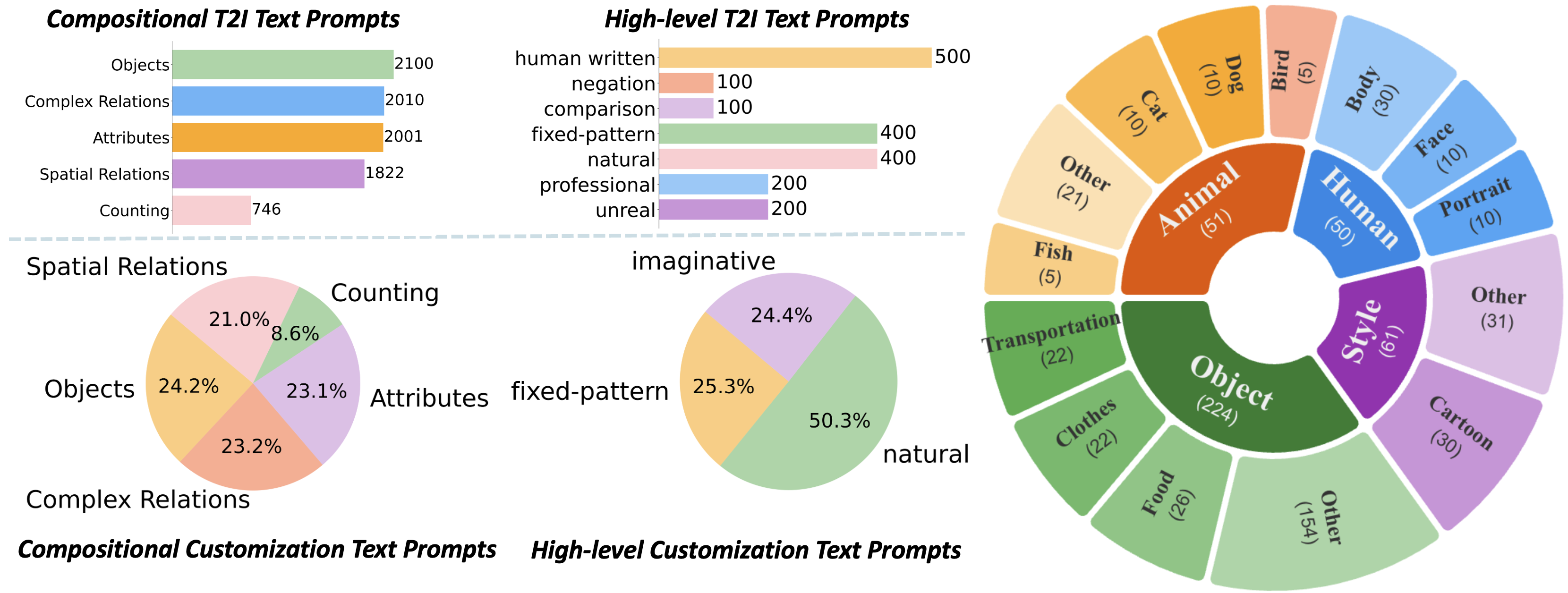}

    \caption{Statistics of the tags in \NAME. \textit{Top-left}: Data distribution of compositional categories and high-level categories for text in T2I task. \textit{Bottom-left}: Data distribution of text prompts in customization task. \textit{Right}: Statistics of classes for the reference images.}
    \label{fig:stat}
\vspace{-4mm}
\end{figure}
\section{Data Curation}
\label{sec:benchmark}

\subsection{Overview}
\label{sec:benchmark_overview}

Multi-modal image generation commonly involves both reference images and text prompts as inputs. Accordingly, our benchmark’s data collection is structured into two components: grouped image collection and text prompt generation (as shown in Fig.~\ref{fig:data_curation}).
We begin by extracting entities from prompts used in existing text-to-image (T2I) benchmarks (such as \cite{Li2024GenAIBenchEA, lee2023holistic, wiles2024revisiting}). After collecting over 2,000 distinct entities, we retain the 207 most frequent entities for subsequent use.

\subsubsection{Prompting GPT for Text Prompt Generation}
\begin{figure}[htbp]
    \centering
    \includegraphics[width=\linewidth]{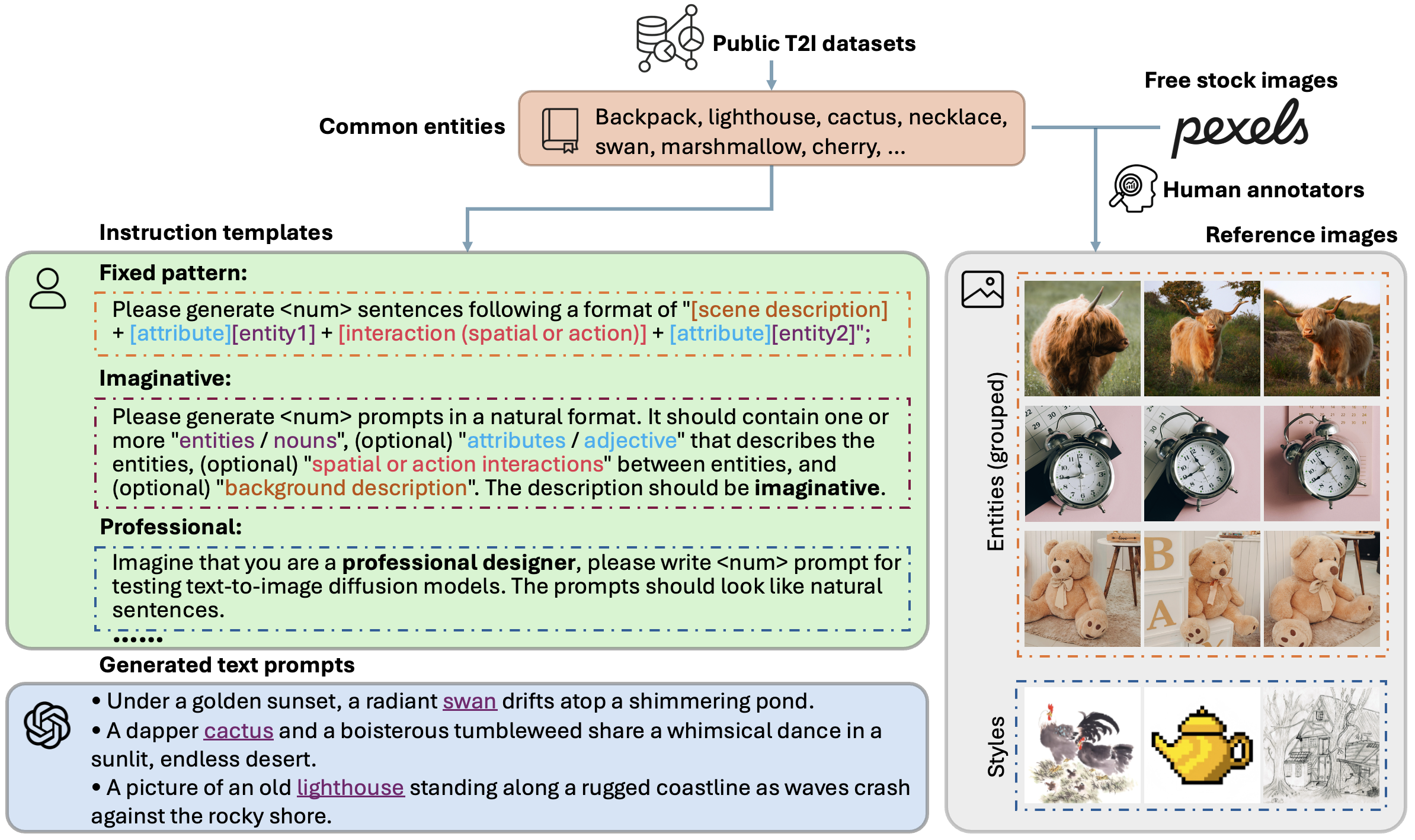}

    \caption{Our data curation pipeline for multi-modal image generation benchmarking. We begin by extracting 207 frequent entities from public T2I datasets. Using these entities, we generate diverse prompts with GPT-4o by prompting it with a set of carefully designed instruction templates, which control the structure and style of the prompts (left). Simultaneously, we collect grouped reference images for each entity from free stock sources, with human annotators selecting 3–5 object-centric images per group that vary in pose or view (right). We further collect artistic images in 12 visual styles to support style transfer. The resulting dataset includes high-quality, structured text-image pairs for both T2I and customization.}
    \label{fig:data_curation}
\vspace{-4mm}
\end{figure}

To enable scalable and diverse prompt generation, we use GPT-4o with several predefined instruction templates, as illustrated in Fig.~\ref{fig:data_curation}. By providing entities and instruction templates as inputs, we generate a total of 4,350 synthetic prompts covering both tasks. Furthermore, we manually select 500 human-written prompts from prior work \cite{Li2024GenAIBenchEA, fu2024commonsense}.
To ensure broad coverage of semantic aspects, we organize prompts into two main categories: compositional and high-level. The compositional category includes five sub-categories: \textit{object}, \textit{counting}, \textit{attribute}, \textit{spatial relations} (e.g., next to, atop, behind), and \textit{complex relations} (e.g., pour into, toss, chase). The high-level category contains seven sub-categories, including \textit{style} (fixed pattern, natural, professional, human-written), 
\textit{realism} (imaginative), and \textit{common sense} (negation, comparison).

To better control the aspects, style, and structure of the prompts, we design eight instruction templates, using the T2I task as an example. When prompts require compositionality and adherence to a specific structure, we use the following format: ``\texttt{[scene description (optional)] + [number][attribute][entity1] + [interaction (spatial or action)] + [number (optional)][attribute][entity2]}''. For prompts to resemble natural, human-written language, a more flexible instruction is used: ``\texttt{Please generate prompts in a NATURAL format. It should contain one or more "entities / nouns", (optional) "attributes / adjective" that describes the entities, (optional) "spatial or action interactions" between entities, and (optional) "background description}''. The full set of templates is provided in the Appendix.

To ensure the quality and safety of the generated prompts, we further filter out toxic or low-quality content (see Sec.~\ref{sec:human_verify}), and utilize FineMatch~\cite{hua2024finematch} to generate dense labels (see Sec.~\ref{sec:parsing}), making the dataset more flexible and suitable for research applications.

\vspace{-3mm}

\subsection{Collecting Grouped Subject Images}
\label{sec:benchmark_images}
Grouped reference images which are object-centric and realistic are usually missing from the previous benchmarks. However, multiple reference images have proven effective across various tasks, including image customization~\cite{ruiz2022dreambooth, kumari2023multi, zong2024easyref}, video generation~\cite{kong2025profashion} and 3D reconstruction~\cite{wang2023pf}. To address this gap, we collect a large set of grouped reference images.

The target objects are selected from the 207 common entities we previously identified. We employ annotators to curate grouped object images from Pexels~\cite{pexels} following these guidelines: (1) each group contains 3–5 images of the same object; (2) within each group, the object appears in varying poses or views; and (3) objects with complex logos or textures are prioritized. Additionally, we collect artistic images in 12 styles (e.g., sketch, low-poly, oil painting) to support style transfer tasks.

In total, we collect 1,750 images across 386 groups, covering four main categories—animals, humans, objects, and styles—as shown in Fig.~\ref{fig:stat} (right). To ensure quality, we apply filtering and cropping to remove unrelated content from the images. Based on the entities in the collected images, we generate corresponding text prompts using the aforementioned procedure.

\subsection{Data Curation for Mid-Level Evaluation}
\label{sec:data_curation_mid}
The goal of mid-level evaluation is to analyze the text-image alignment in fine-grained aspects, enabling more interpretable assessment on the generated details. To this end, we follow FineMatch \cite{hua2024finematch} to analyze the fine-grained text-image alignment from the perspective of \textbf{Object}, \textbf{Relation}, \textbf{Attribute}, and \textbf{Counting}. We conduct specific data curation for these aspects by first using GPT-4o to extract all the aspect-related phrases from input prompts and then using in-context learning to prompt GPT-4o to generate the corresponding QA pairs. 

\subsubsection{Prompt Parsing}
\label{sec:parsing}
We follow FineMatch \cite{hua2024finematch} to curate aspect phrases from text prompts, employing GPT-4o for aspect graph parsing due to its superior compositional parsing capabilities. Specifically, GPT-4o is guided by explicit instructions and in-context examples to accurately extract and categorize phrases into four categories: objects, relations, attributes, and counting queries.

\subsubsection{QA Pair Generation}
Following the prior VQA-based evaluation frameworks \cite{Yarom2023WhatYS,Hu2023TIFAAA,Cho2023DavidsonianSG,hu2023tifa,tang2024vidcomposition,hu2022promptcap,lin2024evaluating,hua2024mmcomposition}, we proceed to generate high-quality question-answer (QA) pairs corresponding to each aspect phrase. Initially, domain experts manually curate a set of exemplar QA pairs for each category (Object, Relation, Attribute, Counting). These manually curated QA pairs serve as contextual examples in the subsequent in-context learning phase. GPT-4o is then prompted with these examples to generate a comprehensive set of QA pairs for the extracted aspect phrases, ensuring alignment with the fine-grained evaluation dimensions. This automated generation process is iteratively refined by adjusting instructions and examples based on preliminary outputs to improve coverage, clarity, and consistency.

\subsection{Human Verification}
\label{sec:human_verify}

To guarantee dataset quality, interpretability, and reliability, we engage trained human annotators in a structured verification process. Annotators perform multiple quality assurance tasks, including: \ding{182} \textbf{Toxicity and Appropriateness Filtering}: Annotators screen generated QA pairs for toxic, offensive, or inappropriate content, ensuring ethical compliance and usability in research settings.\ding{183} \textbf{QA Pair Correction and Validation}: Each QA pair generated by GPT-4o undergoes meticulous human validation to confirm the logical coherence, accuracy, and relevance to the original aspect phrase. Annotators refine ambiguous questions, corrected factual inaccuracies, and ensure precise correspondence between questions and answers. \ding{184} \textbf{Aspect Phrase Refinement}: Extract aspect phrases were scrutinized and refined for linguistic clarity and semantic precision. Annotators review each phrase to ensure they correctly and comprehensively represent the intended compositional aspects (Object, Relation, Attribute, Counting).

After these rigorous human verification steps, we obtain a high-quality dataset consisting of 28,668 (16,819 for T2I tasks and 11,849 for Customization tasks) validated QA pairs, explicitly designed to support detailed analyses of fine-grained text-image alignment. 
\vspace{-10pt}

\section{Proposed Metrics - \NAME}
\label{sec:metrics}

\subsection{Low-Level Evaluation Metrics}
\label{sec:metrics_low}
The goal of low-level evaluation is to assess artifacts in the generated images and to evaluate the low-level feature similarity between the generated images and the prompt, as well as between the generated images and the reference images. To achieve this, we leverage previous evaluation metrics:
\begin{itemize}[left=0pt]
    \item CLIP-Text \cite{Radford2021LearningTV}: measures the semantic alignment between the generated image and input prompt;
    \item CLIP-Image, DINOv2 \cite{Oquab2023DINOv2LR}, and CUTE \cite{Kotar2023AreTT}: measures identity preservation;
    \item PAL4VST \cite{Zhang_2023_ICCV}: measures the amount of generative artifacts using a segmentation model.
\end{itemize}
These metrics collectively provide a comprehensive assessment of the visual quality and consistency.

\vspace{-10pt}
\subsection{Mid-Level Evaluation Metrics}
\label{sec:metrics_mid}
The goal of mid-level evaluation is to assess the fine-grained semantic alignment of generated images with text prompts. We use the collected QA pairs corresponding to the four aspects (as described in Section~\ref{sec:data_curation_mid}) to design a new interpretable evaluation framework, \textbf{Aspect Matching Score (AMS)}.

\vspace{-10pt}
\subsubsection{Aspect Matching Score}
Formally, given a prompt $P$, we extract a set of $n$ aspect phrases $\{A_1, A_2, \dots, A_n\}$ and generate a corresponding set of VQA pairs $\{(Q_1, Ans_1), (Q_2, Ans_2), \dots, (Q_n, Ans_n)\}$. These questions are designed to probe whether the generated image $I$ faithfully reflects the semantics of each aspect.

To compute the alignment score, we use Qwen-VL2.5-72B \cite{bai2025qwen2} to answer each question $Q_i$ based on the generated image $I$, resulting in predicted answers $\{\hat{Ans}_1, \hat{Ans}_2, \dots, \hat{Ans}_n\}$. We then compare each prediction $\hat{Ans}_i$ with the ground truth answer $Ans_i$ to assess correctness. We define the \textbf{Aspect Matching Score}  as the proportion of correctly answered VQA questions:

\vspace{-10pt}
\begin{equation}
\text{AMS}(I, P) = \frac{1}{n} \sum_{i=1}^{n} \mathbf{1}(\hat{Ans}_i = Ans_i),
\end{equation}
\vspace{-10pt}

where $\mathbf{1}(\cdot)$ is an indicator function that returns 1 if the predicted answer exactly matches the ground truth and 0 otherwise. 

\textbf{AMS} provides a direct and interpretable measure of how well the generated image aligns with each semantic component of the prompt. A higher \textbf{AMS} indicates better fine-grained alignment, capturing failures that coarse-level metrics often miss.

\subsection{High-Level Evaluation Metrics}
The goal of high-level evaluation is to evaluate image aesthetics and human preference in the generated images. To achieve this, we leverage previous evaluation metrics, such as Aesthetic, HPSv2 and PickScore. These metrics offer a comprehensive assessment of the visual appeal and alignment with human preferences in the generated outputs.
\section{Experiments}
\label{sec:exp}

\begin{table*}[htbp]
\centering
\caption{Quantitative comparison is conducted across images generated by 12 different text-to-image models using 2,100 well-designed prompts. Most models generate images at the default resolution of 1024 $\times$ 1024, except for the two autoregressive models, which produce outputs at 384 $\times$ 384, and GPT-4o and Gemini-2.0-Flash produce images with variable, non-fixed resolutions.
$\uparrow$ indicates higher is better and $\downarrow$ indicates lower is better. 
The \textbf{best} and \underline{second-best} results are in bold and underlined, respectively.}
\label{tab:t2i_results}
\resizebox{0.9\linewidth}{!}{%
\begin{tabular}{l|cc|cc|ccc}
\toprule
 & \multicolumn{2}{c|}{\textbf{Low Level}} & \multicolumn{2}{c|}{\textbf{Mid Level}}  & \multicolumn{3}{c}{\textbf{High Level}} \\
 \cline{2-8} \\
 \textbf{Method}  & \makecell[c]{\textbf{CLIP-T} $\uparrow$}  & \makecell[c]{\textbf{PAL4VST} $\downarrow$} & \makecell[c]{\textbf{AMS} $\uparrow$}  & \makecell[c]{\textbf{Human} $\uparrow$}  & \makecell[c]{\textbf{Aesthetic} $\uparrow$}  & \makecell[c]{\textbf{HPSv2} $\uparrow$}  & \makecell[c]{\textbf{PickScore} $\uparrow$} \\
\midrule
\rowcolor[HTML]{e9edf6}
\multicolumn{8}{c}{\textbf{Diffusion Models}} \\
\midrule
SDXL \cite{Podell2023SDXLIL} & 33.529 & 14.340 & 79.08 & 72.29 & 6.337 & 0.277 & 0.120 \\
Photon-v1 \cite{photonv1} & 33.296 & 2.947 & 77.12 & 69.49 & 6.391 & 0.284 & 0.088 \\
Lumina-2 \cite{Qin2025LuminaImage2A}& 33.281 & 15.531 & 84.11 & 73.18 & 6.048 & 0.287 & 0.116 \\
HunyuanDit-v1.2 \cite{Li2024HunyuanDiTAP} & 33.701 & 8.024 & 83.61 & 74.89 & 6.379 & 0.300 & 0.144 \\
Pixart-Sigma-xl2 \cite{Chen2024PixArtWT} & 33.682 & 9.283 & 83.18 & 76.65 & 6.409 & 0.304 & 0.165 \\
Flux.1-dev \cite{blackforest2024flux} & 33.017 & \underline{2.171} & 84.44 & 76.44 & \underline{6.433} & \underline{0.307} & 0.210 \\
SD 3.5-large \cite{Esser2024ScalingRF} & \underline{33.873} & 6.359 & \underline{85.33} & 77.04 & 6.318 & 0.294 & 0.157 \\
HiDream-I1-Full \cite{hidream2024i1} &  \textbf{33.876}& \textbf{1.522} & \textbf{89.65} & \textbf{83.18} & \textbf{6.457} & \textbf{0.321} & \textbf{0.450} \\
\midrule
\rowcolor[HTML]{e9edf6}
\multicolumn{8}{c}{\textbf{Autoregressive Models}} \\
\midrule
JanusFlow \cite{Ma2024JanusFlowHA} &  31.498& 365.663 & 70.25 & 75.69 & 5.221 & 0.209 & 0.031 \\
Janus-Pro-7B \cite{Chen2025JanusProUM} &  33.358& 31.954 & 85.35 & 80.36 & 6.038 & 0.275 & 0.129 \\
\midrule
\rowcolor[HTML]{e9edf6}
\multicolumn{8}{c}{\textbf{API-based Models}} \\
\midrule
Gemini-2.0-Flash \cite{google2025gemini2} &  32.433& 11.053 & 85.35 & \underline{81.98} & 6.102 & 0.275 & 0.110 \\
GPT-4o \cite{openai2023gpt4}&  32.380& 3.497 & 82.57 & 81.02 & 6.719 & 0.279 & \underline{0.263} \\
\bottomrule
\end{tabular}
}

\end{table*}

\begin{figure}[htbp]
    \centering

    \includegraphics[width=\linewidth]{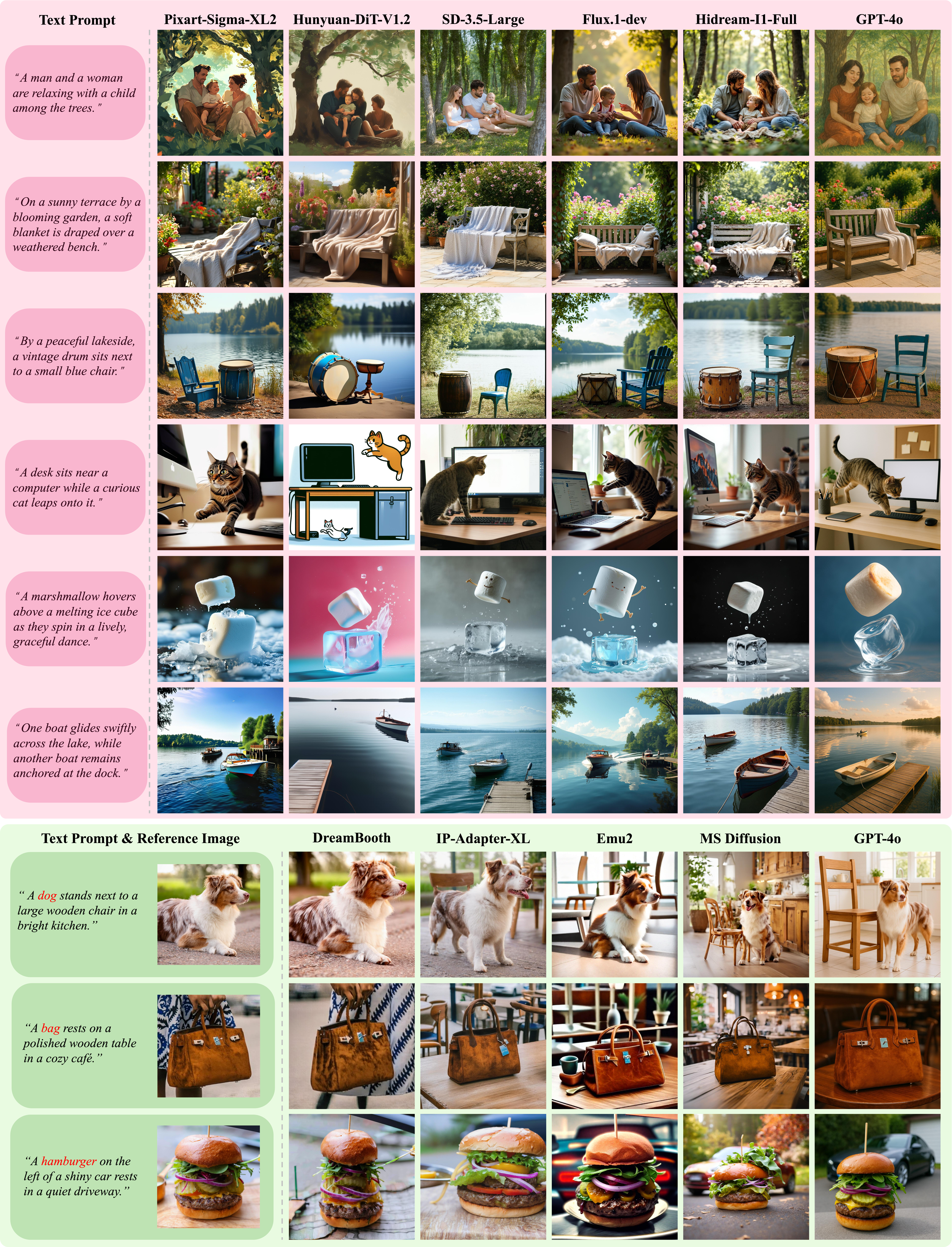}
    \caption{A qualitative study of text-only (top) and text-image-conditioned (bottom) generation methods on \NAME.}
    \label{fig:qual}
\end{figure}

\begin{table*}[htbp]
\centering
\caption{Quantitative comparison is conducted across imagees generated by 6 different multi-modal image generation models using 1,690 samples. Most models generate images 3 times per multi-modal input except GPT-4o at the default resolution of 1024 $\times$ 1024, except for Blip Diffusion, which produce outputs at 512 $\times$ 512, and GPT-4o produce images with variable, non-fixed resolutions. 
$\uparrow$ indicates higher is better and $\downarrow$ indicates lower is better. 
The \textbf{best} and \underline{second best} results are in bold and underlined,  respectively.}
\label{tab:custom_results}
\resizebox{\linewidth}{!}{%
\begin{tabular}{l|ccccc|cc|ccc}
\toprule
 & \multicolumn{5}{c|}{\textbf{Low Level}} & \multicolumn{2}{c|}{\textbf{Mid Level}}  & \multicolumn{3}{c}{\textbf{High Level}} \\
 \cline{2-11} \\
 \textbf{Method}  & \makecell[c]{\textbf{ CLIP-T } $\uparrow$} & \makecell[c]{\textbf{ CLIP-I } $\uparrow$}  & \makecell[c]{\textbf{ DINOv2 } $\uparrow$}  & \makecell[c]{\textbf{ CUTE } $\uparrow$} & \makecell[c]{\textbf{ PAL4VST }$\downarrow$} & \makecell[c]{\textbf{ BLIPVQA }$\uparrow$}  & \makecell[c]{\textbf{ AMS }$\uparrow$} & \makecell[c]{\textbf{ Aesthetic }$\uparrow$}  & \makecell[c]{\textbf{ HPSv2 }$\uparrow$}  & \makecell[c]{\textbf{ PickScore }$\uparrow$}\\
\midrule
\rowcolor[HTML]{e9edf6}
\multicolumn{11}{c}{\textbf{Diffusion Models}} \\
\midrule
BLIP Diffusion\cite{li2023blip} & 26.137 & 80.286 & 26.232 & 69.681 & 56.780 & 0.247 & 41.59 & 5.830 & 0.213 & 0.032 \\
DreamBooth \cite{ruiz2023dreambooth} & 24.227 & \textbf{88.758} & \textbf{38.961} & \textbf{79.780} & 43.535 & 0.108 & 28.00 & 5.368 & 0.179 & 0.019 \\
Emu2 \cite{sun2024generative} & 28.410 & 79.026 & 31.831 & 71.132 & 10.461 & 0.378 & 53.13 & 5.639 & 0.243 & 0.066 \\
Ip-Adapter-XL \cite{ye2023ip} & 28.577 & \underline{85.297} & \underline{34.177} & 74.995 & 8.531 & 0.290 & 51.10 & 5.840 & 0.233 & 0.073 \\
MS Diffusion \cite{wang2024ms} & \underline{31.446} & 77.827 & 23.600 & 71.306 & \underline{4.748} & \underline{0.496} & \underline{71.40} & \underline{5.979} & \underline{0.271} & \underline{0.143} \\
\midrule
\rowcolor[HTML]{e9edf6}
\multicolumn{11}{c}{\textbf{API-based Models}} \\
\midrule
GPT-4o \cite{openai2023gpt4}& \textbf{33.527} & 75.152 & 25.174 & 64.776 & \textbf{1.973} & \textbf{0.672} & \textbf{ 90.90} & \textbf{6.368} & \textbf{0.289} & \textbf{0.550} \\
\bottomrule
\end{tabular}
}

\end{table*}

\begin{table}[htbp]

\centering
\caption{Comparison of VQA-based metrics: BLIPVQA \cite{huang2024t2i}, VQ2 \cite{Yarom2023WhatYS}, DSG \cite{Cho2023DavidsonianSG}, and our \colorbox[HTML]{FFF3CD}{AMS}.}
\label{tab:extra_vqa}
\resizebox{0.7\linewidth}{!}{%
\begin{tabular}{l|cccc|c}
\toprule
\textbf{Method} & \textbf{BLIPVQA} $\uparrow$ & \textbf{VQ2} $\uparrow$ & \textbf{DSG} $\uparrow$ & \textbf{AMS} $\uparrow$ & \textbf{Human} $\uparrow$ \\
\midrule
\rowcolor[HTML]{e9edf6}
\multicolumn{6}{c}{\textbf{Diffusion Models}} \\
\midrule
SDXL & 0.433 & 69.07 & 87.63 & \cellcolor[HTML]{FFF3CD}79.08 & 72.29 \\
Photon-v1 & 0.440 & 66.84 & 86.26 & \cellcolor[HTML]{FFF3CD}77.12 & 69.49 \\
Lumina-2 & 0.517 & 72.51 & 90.12 & \cellcolor[HTML]{FFF3CD}84.11 & 73.18 \\
HunyuanDiT-v1.2 & 0.513 & 73.13 & 89.77 & \cellcolor[HTML]{FFF3CD}83.61 & 74.89 \\
Pixart-Sigma-xl2 & 0.521 & 71.51 & 89.69 & \cellcolor[HTML]{FFF3CD}83.18 & 76.65 \\
Flux.1-dev & 0.511 & 71.41 & 83.33 & \cellcolor[HTML]{FFF3CD}84.44 & 76.44 \\
SD 3.5-large & 0.525 & 73.28 & 91.41 & \cellcolor[HTML]{FFF3CD}85.33 & 77.04 \\
HiDream-I1-Full & 0.572 & 75.09 & 92.43 & \cellcolor[HTML]{FFF3CD}89.65 & 83.18 \\
\midrule
\rowcolor[HTML]{e9edf6}
\multicolumn{6}{c}{\textbf{Autoregressive Models}} \\
\midrule
JanusFlow \cite{Ma2024JanusFlowHA} & 0.390 & 57.24 & 85.43 & \cellcolor[HTML]{FFF3CD}70.25 & 75.69 \\
Janus-Pro \cite{Chen2025JanusProUM} & 0.530 & 67.41 & 92.15 & \cellcolor[HTML]{FFF3CD}85.35 & 80.36 \\
\midrule
\rowcolor[HTML]{e9edf6}
\multicolumn{6}{c}{\textbf{API-based Models}} \\
\midrule
Gemini-2.0-Flash & 0.495 & 72.01 & 92.93 & \cellcolor[HTML]{FFF3CD}85.40 & 81.98 \\
GPT-4o & 0.497 & 70.34 & 89.99 & \cellcolor[HTML]{FFF3CD}82.57 & 81.02 \\
\bottomrule
\end{tabular}
}

\end{table}

\subsection{Human Evaluation}
To evaluate the semantic preservation of state-of-the-art generation models and compare the human correlation of VQA-based metrics, we conduct five user studies. We assess 12 text-to-image (T2I) models across five aspects: attribute, relation, counting, object, and general prompt following. For each of the first four aspects, 150 prompts are randomly selected; for the last, 300 prompts are used.
In each study, users are shown a prompt and a generated image, and asked to rate semantic alignment on a 1–5 scale based on the target aspect (see Appendix for details). In total, we collect 32.4k ratings from over 8,000 Amazon Mechanical Turk users. Results are reported in Table~\ref{tab:extra_vqa}.

\subsection{Correlation of Automated Metrics with Human Annotations}
\label{sec:exp_correlation}
To assess the alignment of automated metrics with human, we compute Spearman correlations against human annotations. As shown in Table~\ref{tab:extra_vqa}, our proposed \colorbox[HTML]{FFF3CD}{AMS} achieves the highest correlation ($\rho =$ \textbf{0.699}), surpassing DSG ($\rho = 0.692$), VQ2 ($\rho = 0.399$), and BLIPVQA ($\rho = 0.147$). This demonstrates the effectiveness of AMS as a reliable metric for compositional T2I evaluation.

\subsection{Leaderboard}
\label{sec:exp_leaderboard}

We compare the performance across state-of-the-art models in T2I task (Tab.~\ref{tab:t2i_results}) and customization task (Tab.~\ref{tab:custom_results}) using our multi-level evaluation framework. Based on the scores, we can derive the following insights:

In T2I task: (1) Compared with diffusion models, autoregressive models (JanusFlow and Janus-Pro-7B) perform significantly worse in visual quality, as they are more likely to generate artifacts, and have the lowest aesthetic and human preference scores. (2) HiDream-I1, the largest model with 17B parameters, excels all the other generators; it takes advantage of rectified flow and the VAE from FLUX.1-schnell. (3) FLUX.1-dev (the second largest model with 12B parameters) stands at the second place for most metrics. (4) The performance of HiDream-I1 and FLUX.1-dev suggests the importance of scaling  generative models. (5) Although GPT-4o is not the best model in all metrics, it shows very robust generation abilities competitive to the best model in each category.

In customization task, we draw the following conclusions: (1) In most low-level metrics that evaluates identity preservation, DreamBooth is the strongest model; its multi-view inputs and test-time finetuning greatly enhances the identity learning. (2) GPT-4o cannot preserve the identity well, this ability is even worse than some early models like Emu2 and the two encoder-based models (BLIP Diffusion and IP-Adapter). (3) GPT-4o comes at the first place in visual quality and semantic alignment. (4) MS Diffusion is often the second best in terms of generation quality, validating the effectiveness of the grounding resampler and MS cross-attention. However, it shows an unsatisfactory ability on identity preservation.

\subsection{Qualitative Analysis}
\label{sec:exp_qual}

We present qualitative results for multi-modal image generation in Fig.~\ref{fig:qual}. The top six rows illustrate generations conditioned on text only; the bottom three rows show generations conditioned on both image and text. Key observations are as follows:

In the T2I task, (1) Hunyuan-DiT-V1.2 struggles with entity generation, frequently missing objects, duplicating them, or generating incorrect ones; (2) Pixart-Sigma-XL2 exhibits stronger visual artifacts (e.g., around benches, chairs, and computers), consistent with its lower PAL4VST scores from Tab.~\ref{tab:t2i_results}. In customization task, (1) Non-rigid objects (e.g., dogs) tend to appear in more diverse poses; (2) MS-Diffusion performs worst in preserving object identity, while DreamBooth performs best; This highly aligns with the CLIP-I and DINOv2 scores in Tab.~\ref{tab:custom_results}. (3) Despite its strength in identity preservation, DreamBooth often fails to generate the correct scene, actions, or additional entities, indicating poor compositional alignment.

\section{Discussions and Conclusions}
\label{sec:conclusion}

We present \NAME, the first benchmark to treat multi-modal image generation as a single task rather than two disjoint tasks. We demonstrate that by pairing 1,750 multi-view reference images with 4,850 densely annotated prompts, \NAME enables side-by-side evaluation of pure text-to-image, image-conditioned customization, and every hybrid in between. The proposed three-level evaluation framework provides a comprehensive, interpretable assessment that addresses the evaluation gaps in both T2I and customization tasks. The evaluation metrics prove to be well aligned with human preferences by comparing them with 32k human ratings across 17 state-of-the-art models. The in-depth assessments of the image generators on our benchmark provide insights on how the model capacity, model architecture, and other factors influence the image quality. One limitation is that the human ratings do not yet cover visual quality; we plan to expand future studies to such dimensions. We will publicly release the data, code, and leaderboard to encourage transparent comparison and guide future advances in architecture design, data curation, and training strategy.

\bibliography{main}

\begin{thebibliography}{65}
\providecommand{\natexlab}[1]{#1}
\providecommand{\url}[1]{\texttt{#1}}
\expandafter\ifx\csname urlstyle\endcsname\relax
  \providecommand{\doi}[1]{doi: #1}\else
  \providecommand{\doi}{doi: \begingroup \urlstyle{rm}\Url}\fi

\bibitem[Bai et~al.(2025)Bai, Chen, Liu, Wang, Ge, Song, Dang, Wang, Wang, Tang, et~al.]{bai2025qwen2}
Shuai Bai, Keqin Chen, Xuejing Liu, Jialin Wang, Wenbin Ge, Sibo Song, Kai Dang, Peng Wang, Shijie Wang, Jun Tang, et~al.
\newblock Qwen2. 5-vl technical report.
\newblock \emph{arXiv preprint arXiv:2502.13923}, 2025.

\bibitem[Chen et~al.(2024)Chen, Ge, Xie, Wu, Yao, Ren, Wang, Luo, Lu, and Li]{Chen2024PixArtWT}
Junsong Chen, Chongjian Ge, Enze Xie, Yue Wu, Lewei Yao, Xiaozhe Ren, Zhongdao Wang, Ping Luo, Huchuan Lu, and Zhenguo Li.
\newblock Pixart-$\sigma$: Weak-to-strong training of diffusion transformer for 4k text-to-image generation.
\newblock In \emph{European Conference on Computer Vision}, 2024.
\newblock URL \url{https://api.semanticscholar.org/CorpusID:268264262}.

\bibitem[Chen et~al.(2025)Chen, Wu, Liu, Pan, Liu, Xie, Yu, and Ruan]{Chen2025JanusProUM}
Xiaokang Chen, Zhiyu Wu, Xingchao Liu, Zizheng Pan, Wen Liu, Zhenda Xie, Xingkai Yu, and Chong Ruan.
\newblock Janus-pro: Unified multimodal understanding and generation with data and model scaling.
\newblock \emph{ArXiv}, abs/2501.17811, 2025.
\newblock URL \url{https://api.semanticscholar.org/CorpusID:275954151}.

\bibitem[Cho et~al.(2023)Cho, Hu, Garg, Anderson, Krishna, Baldridge, Bansal, Pont-Tuset, and Wang]{Cho2023DavidsonianSG}
Jaemin Cho, Yushi Hu, Roopal Garg, Peter Anderson, Ranjay Krishna, Jason Baldridge, Mohit Bansal, Jordi Pont-Tuset, and Su~Wang.
\newblock Davidsonian scene graph: Improving reliability in fine-grained evaluation for text-to-image generation.
\newblock \emph{ArXiv}, abs/2310.18235, 2023.
\newblock URL \url{https://api.semanticscholar.org/CorpusID:264555374}.

\bibitem[dreambench(2022)]{dreambench}
dreambench.
\newblock Dreambench, 2022.
\newblock \url{https://github.com/nousr/dream-bench}.

\bibitem[Esser et~al.(2024)Esser, Kulal, Blattmann, Entezari, Muller, Saini, Levi, Lorenz, Sauer, Boesel, Podell, Dockhorn, English, Lacey, Goodwin, Marek, and Rombach]{Esser2024ScalingRF}
Patrick Esser, Sumith Kulal, A.~Blattmann, Rahim Entezari, Jonas Muller, Harry Saini, Yam Levi, Dominik Lorenz, Axel Sauer, Frederic Boesel, Dustin Podell, Tim Dockhorn, Zion English, Kyle Lacey, Alex Goodwin, Yannik Marek, and Robin Rombach.
\newblock Scaling rectified flow transformers for high-resolution image synthesis.
\newblock \emph{ArXiv}, abs/2403.03206, 2024.
\newblock URL \url{https://api.semanticscholar.org/CorpusID:268247980}.

\bibitem[Feng et~al.(2025)Feng, Huang, Li, Lv, and Sheng]{feng2025personalize}
Haoran Feng, Zehuan Huang, Lin Li, Hairong Lv, and Lu~Sheng.
\newblock Personalize anything for free with diffusion transformer.
\newblock \emph{arXiv preprint arXiv:2503.12590}, 2025.

\bibitem[Fu et~al.(2024)Fu, He, Lu, Wang, and Roth]{fu2024commonsense}
Xingyu Fu, Muyu He, Yujie Lu, William~Yang Wang, and Dan Roth.
\newblock Commonsense-t2i challenge: Can text-to-image generation models understand commonsense?
\newblock \emph{arXiv preprint arXiv:2406.07546}, 2024.

\bibitem[Gal et~al.(2022)Gal, Alaluf, Atzmon, Patashnik, Bermano, Chechik, and Cohen-Or]{gal2022image}
Rinon Gal, Yuval Alaluf, Yuval Atzmon, Or~Patashnik, Amit~H Bermano, Gal Chechik, and Daniel Cohen-Or.
\newblock An image is worth one word: Personalizing text-to-image generation using textual inversion.
\newblock \emph{arXiv preprint arXiv:2208.01618}, 2022.

\bibitem[Ghosh et~al.(2023)Ghosh, Hajishirzi, and Schmidt]{ghosh2023geneval}
Dhruba Ghosh, Hannaneh Hajishirzi, and Ludwig Schmidt.
\newblock Geneval: An object-focused framework for evaluating text-to-image alignment.
\newblock \emph{Advances in Neural Information Processing Systems}, 36:\penalty0 52132--52152, 2023.

\bibitem[Google(2025)]{google2025gemini2}
Google.
\newblock Gemini 2.0 flash, 2025.
\newblock \url{https://blog.google/technology/google-deepmind/google-gemini-ai-update-december-2024/#gemini-2-0-flash}.

\bibitem[Han et~al.(2024)Han, Fan, Fu, Li, Li, Cui, Wang, Tai, Sun, Guo, and Li]{Han2024EvalMuse40KAR}
Shuhao Han, Haotian Fan, Jiachen Fu, Liang Li, Tao Li, Junhui Cui, Yunqiu Wang, Yang Tai, Jingwei Sun, Chunle Guo, and Chongyi Li.
\newblock Evalmuse-40k: A reliable and fine-grained benchmark with comprehensive human annotations for text-to-image generation model evaluation.
\newblock \emph{ArXiv}, abs/2412.18150, 2024.
\newblock URL \url{https://api.semanticscholar.org/CorpusID:274992412}.

\bibitem[Hu et~al.(2024)Hu, Wang, Fang, Fu, Cheng, and Yu]{hu2024ella}
Xiwei Hu, Rui Wang, Yixiao Fang, Bin Fu, Pei Cheng, and Gang Yu.
\newblock Ella: Equip diffusion models with llm for enhanced semantic alignment.
\newblock \emph{arXiv preprint arXiv:2403.05135}, 2024.

\bibitem[Hu et~al.(2022)Hu, Hua, Yang, Shi, Smith, and Luo]{hu2022promptcap}
Yushi Hu, Hang Hua, Zhengyuan Yang, Weijia Shi, Noah~A Smith, and Jiebo Luo.
\newblock Promptcap: Prompt-guided task-aware image captioning.
\newblock \emph{arXiv preprint arXiv:2211.09699}, 2022.

\bibitem[Hu et~al.(2023{\natexlab{a}})Hu, Liu, Kasai, Wang, Ostendorf, Krishna, and Smith]{Hu2023TIFAAA}
Yushi Hu, Benlin Liu, Jungo Kasai, Yizhong Wang, Mari Ostendorf, Ranjay Krishna, and Noah~A. Smith.
\newblock Tifa: Accurate and interpretable text-to-image faithfulness evaluation with question answering.
\newblock \emph{2023 IEEE/CVF International Conference on Computer Vision (ICCV)}, pages 20349--20360, 2023{\natexlab{a}}.
\newblock URL \url{https://api.semanticscholar.org/CorpusID:257636562}.

\bibitem[Hu et~al.(2023{\natexlab{b}})Hu, Liu, Kasai, Wang, Ostendorf, Krishna, and Smith]{hu2023tifa}
Yushi Hu, Benlin Liu, Jungo Kasai, Yizhong Wang, Mari Ostendorf, Ranjay Krishna, and Noah~A Smith.
\newblock Tifa: Accurate and interpretable text-to-image faithfulness evaluation with question answering.
\newblock In \emph{Proceedings of the IEEE/CVF International Conference on Computer Vision}, pages 20406--20417, 2023{\natexlab{b}}.

\bibitem[Hua et~al.(2024{\natexlab{a}})Hua, Shi, Kafle, Jenni, Zhang, Collomosse, Cohen, and Luo]{hua2024finematch}
Hang Hua, Jing Shi, Kushal Kafle, Simon Jenni, Daoan Zhang, John Collomosse, Scott Cohen, and Jiebo Luo.
\newblock Finematch: Aspect-based fine-grained image and text mismatch detection and correction.
\newblock In \emph{European Conference on Computer Vision}, pages 474--491. Springer, 2024{\natexlab{a}}.

\bibitem[Hua et~al.(2024{\natexlab{b}})Hua, Tang, Zeng, Cao, Yang, He, Xu, and Luo]{hua2024mmcomposition}
Hang Hua, Yunlong Tang, Ziyun Zeng, Liangliang Cao, Zhengyuan Yang, Hangfeng He, Chenliang Xu, and Jiebo Luo.
\newblock Mmcomposition: Revisiting the compositionality of pre-trained vision-language models.
\newblock \emph{arXiv preprint arXiv:2410.09733}, 2024{\natexlab{b}}.

\bibitem[Huang et~al.(2023)Huang, Duan, Sun, Xie, Li, and Liu]{Huang2023T2ICompBenchAE}
Kaiyi Huang, Chengqi Duan, Kaiyue Sun, Enze Xie, Zhenguo Li, and Xihui Liu.
\newblock T2i-compbench++: An enhanced and comprehensive benchmark for compositional text-to-image generation.
\newblock \emph{IEEE Transactions on Pattern Analysis and Machine Intelligence}, 47:\penalty0 3563--3579, 2023.
\newblock URL \url{https://api.semanticscholar.org/CorpusID:259847295}.

\bibitem[Huang et~al.(2024)Huang, He, Long, Wang, Li, Yu, Shu, Chan, Jiang, Gan, et~al.]{huang2024t2i}
Ziwei Huang, Wanggui He, Quanyu Long, Yandi Wang, Haoyuan Li, Zhelun Yu, Fangxun Shu, Long Chan, Hao Jiang, Leilei Gan, et~al.
\newblock T2i-factualbench: Benchmarking the factuality of text-to-image models with knowledge-intensive concepts.
\newblock \emph{arXiv preprint arXiv:2412.04300}, 2024.

\bibitem[Kawar et~al.(2023)Kawar, Zada, Lang, Tov, Chang, Dekel, Mosseri, and Irani]{kawar2023imagic}
Bahjat Kawar, Shiran Zada, Oran Lang, Omer Tov, Huiwen Chang, Tali Dekel, Inbar Mosseri, and Michal Irani.
\newblock Imagic: Text-based real image editing with diffusion models.
\newblock In \emph{Proceedings of the IEEE/CVF conference on computer vision and pattern recognition}, pages 6007--6017, 2023.

\bibitem[Kong et~al.(2025)Kong, Qi, Wang, Rao, Chen, Zhang, Liu, and Jiang]{kong2025profashion}
Xianghao Kong, Qiaosong Qi, Yuanbin Wang, Anyi Rao, Biaolong Chen, Aixi Zhang, Si~Liu, and Hao Jiang.
\newblock Profashion: Prototype-guided fashion video generation with multiple reference images.
\newblock \emph{arXiv preprint arXiv:2505.06537}, 2025.

\bibitem[Kotar et~al.(2023)Kotar, Tian, Yu, Yamins, and Wu]{Kotar2023AreTT}
Klemen Kotar, Stephen Tian, Hong-Xing Yu, Daniel L.~K. Yamins, and Jiajun Wu.
\newblock Are these the same apple? comparing images based on object intrinsics.
\newblock \emph{ArXiv}, abs/2311.00750, 2023.
\newblock URL \url{https://api.semanticscholar.org/CorpusID:264935263}.

\bibitem[Kumari et~al.(2023)Kumari, Zhang, Zhang, Shechtman, and Zhu]{kumari2023multi}
Nupur Kumari, Bingliang Zhang, Richard Zhang, Eli Shechtman, and Jun-Yan Zhu.
\newblock Multi-concept customization of text-to-image diffusion.
\newblock In \emph{Proceedings of the IEEE/CVF conference on computer vision and pattern recognition}, pages 1931--1941, 2023.

\bibitem[Labs(2024{\natexlab{a}})]{blackforest2024flux}
Black~Forest Labs.
\newblock Flux.1, 2024{\natexlab{a}}.
\newblock \url{https://bfl.ai/announcements/24-08-01-bfl}.

\bibitem[Labs(2024{\natexlab{b}})]{flux2024}
Black~Forest Labs.
\newblock Flux.
\newblock \url{https://github.com/black-forest-labs/flux}, 2024{\natexlab{b}}.

\bibitem[Lee et~al.(2023)Lee, Yasunaga, Meng, Mai, Park, Gupta, Zhang, Narayanan, Teufel, Bellagente, et~al.]{lee2023holistic}
Tony Lee, Michihiro Yasunaga, Chenlin Meng, Yifan Mai, Joon~Sung Park, Agrim Gupta, Yunzhi Zhang, Deepak Narayanan, Hannah Teufel, Marco Bellagente, et~al.
\newblock Holistic evaluation of text-to-image models.
\newblock \emph{Advances in Neural Information Processing Systems}, 36:\penalty0 69981--70011, 2023.

\bibitem[Li et~al.(2024{\natexlab{a}})Li, Lin, Pathak, Li, Fei, Wu, Ling, Xia, Zhang, Neubig, and Ramanan]{Li2024GenAIBenchEA}
Baiqi Li, Zhiqiu Lin, Deepak Pathak, Jiayao Li, Yixin Fei, Kewen Wu, Tiffany Ling, Xide Xia, Pengchuan Zhang, Graham Neubig, and Deva Ramanan.
\newblock Genai-bench: Evaluating and improving compositional text-to-visual generation.
\newblock \emph{ArXiv}, abs/2406.13743, 2024{\natexlab{a}}.
\newblock URL \url{https://api.semanticscholar.org/CorpusID:270619531}.

\bibitem[Li et~al.(2023)Li, Li, and Hoi]{li2023blip}
Dongxu Li, Junnan Li, and Steven Hoi.
\newblock Blip-diffusion: Pre-trained subject representation for controllable text-to-image generation and editing.
\newblock \emph{Advances in Neural Information Processing Systems}, 36:\penalty0 30146--30166, 2023.

\bibitem[Li et~al.(2024{\natexlab{b}})Li, Xu, Liu, and Xiao]{li2024unimo}
Wei Li, Xue Xu, Jiachen Liu, and Xinyan Xiao.
\newblock Unimo-g: Unified image generation through multimodal conditional diffusion.
\newblock \emph{arXiv preprint arXiv:2401.13388}, 2024{\natexlab{b}}.

\bibitem[Li et~al.(2024{\natexlab{c}})Li, Zhang, Lin, Xiong, Long, Deng, Zhang, Liu, Huang, Xiao, Chen, He, Li, Li, Zhang, Quan, Lu, Huang, Yuan, Zheng, Li, Zhang, Zhang, Chen, Liu, Fang, Wang, Xue, Tao, Zhu, Liu, Lin, Sun, Li, Wang, Chen, Hu, Xiao, Chen, Liu, Liu, Wang, Yang, Jiang, and Lu]{Li2024HunyuanDiTAP}
Zhimin Li, Jianwei Zhang, Qin Lin, Jiangfeng Xiong, Yanxin Long, Xinchi Deng, Yingfang Zhang, Xingchao Liu, Minbin Huang, Zedong Xiao, Dayou Chen, Jiajun He, Jiahao Li, Wenyue Li, Chen Zhang, Rongwei Quan, Jianxiang Lu, Jiabin Huang, Xiaoyan Yuan, Xiao-Ting Zheng, Yixuan Li, Jihong Zhang, Chao Zhang, Mengxi Chen, Jie Liu, Zheng Fang, Weiyan Wang, Jinbao Xue, Yang-Dan Tao, Jianchen Zhu, Kai Liu, Si-Da Lin, Yifu Sun, Yun Li, Dongdong Wang, Mingtao Chen, Zhichao Hu, Xiao Xiao, Yan Chen, Yuhong Liu, Wei Liu, Dingyong Wang, Yong Yang, Jie Jiang, and Qinglin Lu.
\newblock Hunyuan-dit: A powerful multi-resolution diffusion transformer with fine-grained chinese understanding.
\newblock \emph{ArXiv}, abs/2405.08748, 2024{\natexlab{c}}.
\newblock URL \url{https://api.semanticscholar.org/CorpusID:269761491}.

\bibitem[Lin et~al.(2024)Lin, Pathak, Li, Li, Xia, Neubig, Zhang, and Ramanan]{lin2024evaluating}
Zhiqiu Lin, Deepak Pathak, Baiqi Li, Jiayao Li, Xide Xia, Graham Neubig, Pengchuan Zhang, and Deva Ramanan.
\newblock Evaluating text-to-visual generation with image-to-text generation.
\newblock In \emph{European Conference on Computer Vision}, pages 366--384. Springer, 2024.

\bibitem[Ma et~al.(2024)Ma, Liu, Chen, Liu, Wu, Wu, Pan, Xie, Zhang, Yu, Zhao, Wang, Liu, and Ruan]{Ma2024JanusFlowHA}
Yiyang Ma, Xingchao Liu, Xiaokang Chen, Wen Liu, Chengyue Wu, Zhiyu Wu, Zizheng Pan, Zhenda Xie, Haowei Zhang, Xingkai Yu, Liang Zhao, Yisong Wang, Jiaying Liu, and Chong Ruan.
\newblock Janusflow: Harmonizing autoregression and rectified flow for unified multimodal understanding and generation.
\newblock \emph{ArXiv}, abs/2411.07975, 2024.
\newblock URL \url{https://api.semanticscholar.org/CorpusID:273969525}.

\bibitem[Mou et~al.(2025)Mou, Wu, Wu, Guo, Zhang, Cheng, Luo, Ding, Zhang, Li, et~al.]{mou2025dreamo}
Chong Mou, Yanze Wu, Wenxu Wu, Zinan Guo, Pengze Zhang, Yufeng Cheng, Yiming Luo, Fei Ding, Shiwen Zhang, Xinghui Li, et~al.
\newblock Dreamo: A unified framework for image customization.
\newblock \emph{arXiv preprint arXiv:2504.16915}, 2025.

\bibitem[OpenAI(2023)]{openai2023gpt4}
OpenAI.
\newblock Gpt-4 technical report.
\newblock \emph{arXiv preprint arXiv:2303.08774}, 2023.
\newblock URL \url{https://arxiv.org/abs/2303.08774}.

\bibitem[Oquab et~al.(2023)Oquab, Darcet, Moutakanni, Vo, Szafraniec, Khalidov, Fernandez, Haziza, Massa, El-Nouby, Assran, Ballas, Galuba, Howes, Huang, Li, Misra, Rabbat, Sharma, Synnaeve, Xu, J{\'e}gou, Mairal, Labatut, Joulin, and Bojanowski]{Oquab2023DINOv2LR}
Maxime Oquab, Timoth{\'e}e Darcet, Th{\'e}o Moutakanni, Huy~Q. Vo, Marc Szafraniec, Vasil Khalidov, Pierre Fernandez, Daniel Haziza, Francisco Massa, Alaaeldin El-Nouby, Mahmoud Assran, Nicolas Ballas, Wojciech Galuba, Russ Howes, Po-Yao~(Bernie) Huang, Shang-Wen Li, Ishan Misra, Michael~G. Rabbat, Vasu Sharma, Gabriel Synnaeve, Huijiao Xu, Herv{\'e} J{\'e}gou, Julien Mairal, Patrick Labatut, Armand Joulin, and Piotr Bojanowski.
\newblock Dinov2: Learning robust visual features without supervision.
\newblock \emph{ArXiv}, abs/2304.07193, 2023.
\newblock URL \url{https://api.semanticscholar.org/CorpusID:258170077}.

\bibitem[Pan et~al.(2023)Pan, Dong, Huang, Peng, Chen, and Wei]{Pan2023KosmosGGI}
Xichen Pan, Li~Dong, Shaohan Huang, Zhiliang Peng, Wenhu Chen, and Furu Wei.
\newblock Kosmos-g: Generating images in context with multimodal large language models.
\newblock \emph{ArXiv}, abs/2310.02992, 2023.
\newblock URL \url{https://api.semanticscholar.org/CorpusID:263620748}.

\bibitem[Peng et~al.(2024)Peng, Cui, Tang, Qi, Dong, Bai, Han, Ge, Zhang, and Xia]{Peng2024DreamBenchAH}
Yuang Peng, Yuxin Cui, Haomiao Tang, Zekun Qi, Runpei Dong, Jing Bai, Chunrui Han, Zheng Ge, Xiangyu Zhang, and Shu-Tao Xia.
\newblock Dreambench++: A human-aligned benchmark for personalized image generation.
\newblock \emph{ArXiv}, abs/2406.16855, 2024.
\newblock URL \url{https://api.semanticscholar.org/CorpusID:270702690}.

\bibitem[Pexels(2014)]{pexels}
Pexels.
\newblock Pexels, 2014.
\newblock \url{https://www.pexels.com/}.

\bibitem[Photon78(2023)]{photonv1}
Photon78.
\newblock Photon-v1.
\newblock \url{https://civitai.com/models/84728/photon78}, 2023.
\newblock Accessed: 2025-05-06.

\bibitem[Podell et~al.(2023)Podell, English, Lacey, Blattmann, Dockhorn, Muller, Penna, and Rombach]{Podell2023SDXLIL}
Dustin Podell, Zion English, Kyle Lacey, A.~Blattmann, Tim Dockhorn, Jonas Muller, Joe Penna, and Robin Rombach.
\newblock Sdxl: Improving latent diffusion models for high-resolution image synthesis.
\newblock \emph{ArXiv}, abs/2307.01952, 2023.
\newblock URL \url{https://api.semanticscholar.org/CorpusID:259341735}.

\bibitem[Qin et~al.(2025)Qin, Zhuo, Xin, Du, Li, Fu, Lu, Yuan, Li, Liu, Zhu, Zhang, Beddow, Millon, Perez, Wang, He, Zhang, Liu, Li, Qiao, Xu, and Gao]{Qin2025LuminaImage2A}
Qi~Qin, Le~Zhuo, Yi~Xin, Ruoyi Du, Zhen Li, Bin Fu, Yiting Lu, Jiakang Yuan, Xinyue Li, Dongyang Liu, Xiangyang Zhu, Manyuan Zhang, Will Beddow, Erwann Millon, Victor Perez, Wen-Hao Wang, Conghui He, Bo~Zhang, Xiaohong Liu, Hongsheng Li, Yu-Hao Qiao, Chang Xu, and Peng Gao.
\newblock Lumina-image 2.0: A unified and efficient image generative framework.
\newblock \emph{ArXiv}, abs/2503.21758, 2025.
\newblock URL \url{https://api.semanticscholar.org/CorpusID:277349538}.

\bibitem[Radford et~al.(2021)Radford, Kim, Hallacy, Ramesh, Goh, Agarwal, Sastry, Askell, Mishkin, Clark, Krueger, and Sutskever]{Radford2021LearningTV}
Alec Radford, Jong~Wook Kim, Chris Hallacy, Aditya Ramesh, Gabriel Goh, Sandhini Agarwal, Girish Sastry, Amanda Askell, Pamela Mishkin, Jack Clark, Gretchen Krueger, and Ilya Sutskever.
\newblock Learning transferable visual models from natural language supervision.
\newblock In \emph{International Conference on Machine Learning}, 2021.
\newblock URL \url{https://api.semanticscholar.org/CorpusID:231591445}.

\bibitem[Ruiz et~al.(2022)Ruiz, Li, Jampani, Pritch, Rubinstein, and Aberman]{ruiz2022dreambooth}
Nataniel Ruiz, Yuanzhen Li, Varun Jampani, Yael Pritch, Michael Rubinstein, and Kfir Aberman.
\newblock Dreambooth: Fine tuning text-to-image diffusion models for subject-driven generation.
\newblock 2022.

\bibitem[Ruiz et~al.(2023{\natexlab{a}})Ruiz, Li, Jampani, Pritch, Rubinstein, and Aberman]{ruiz2023dreambooth}
Nataniel Ruiz, Yuanzhen Li, Varun Jampani, Yael Pritch, Michael Rubinstein, and Kfir Aberman.
\newblock Dreambooth: Fine tuning text-to-image diffusion models for subject-driven generation.
\newblock In \emph{Proceedings of the IEEE/CVF conference on computer vision and pattern recognition}, pages 22500--22510, 2023{\natexlab{a}}.

\bibitem[Ruiz et~al.(2023{\natexlab{b}})Ruiz, Li, Jampani, Wei, Hou, Pritch, Wadhwa, Rubinstein, and Aberman]{ruiz2023hyperdreambooth}
Nataniel Ruiz, Yuanzhen Li, Varun Jampani, Wei Wei, Tingbo Hou, Yael Pritch, Neal Wadhwa, Michael Rubinstein, and Kfir Aberman.
\newblock Hyperdreambooth: Hypernetworks for fast personalization of text-to-image models, 2023{\natexlab{b}}.

\bibitem[Shi et~al.(2024)Shi, Xiong, Lin, and Jung]{shi2024instantbooth}
Jing Shi, Wei Xiong, Zhe Lin, and Hyun~Joon Jung.
\newblock Instantbooth: Personalized text-to-image generation without test-time finetuning.
\newblock In \emph{Proceedings of the IEEE/CVF conference on computer vision and pattern recognition}, pages 8543--8552, 2024.

\bibitem[Sun et~al.(2024)Sun, Cui, Zhang, Zhang, Yu, Wang, Rao, Liu, Huang, and Wang]{sun2024generative}
Quan Sun, Yufeng Cui, Xiaosong Zhang, Fan Zhang, Qiying Yu, Yueze Wang, Yongming Rao, Jingjing Liu, Tiejun Huang, and Xinlong Wang.
\newblock Generative multimodal models are in-context learners.
\newblock In \emph{Proceedings of the IEEE/CVF Conference on Computer Vision and Pattern Recognition}, pages 14398--14409, 2024.

\bibitem[Tang et~al.(2024)Tang, Guo, Hua, Liang, Feng, Li, Mao, Huang, Bi, Zhang, et~al.]{tang2024vidcomposition}
Yunlong Tang, Junjia Guo, Hang Hua, Susan Liang, Mingqian Feng, Xinyang Li, Rui Mao, Chao Huang, Jing Bi, Zeliang Zhang, et~al.
\newblock Vidcomposition: Can mllms analyze compositions in compiled videos?
\newblock \emph{arXiv preprint arXiv:2411.10979}, 2024.

\bibitem[Team(2024)]{hidream2024i1}
HiDream-AI Team.
\newblock Hidream-i1: A 17b parameter open chinese text-to-image generation model.
\newblock \url{https://github.com/HiDream-ai/HiDream-I1}, 2024.
\newblock Accessed: 2025-05-14.

\bibitem[Wang et~al.(2025)Wang, Duan, Zhao, Wang, Zhai, and Min]{Wang2025LMM4LMMBA}
Jiarui Wang, Huiyu Duan, Yu~Zhao, Juntong Wang, Guangtao Zhai, and Xiongkuo Min.
\newblock Lmm4lmm: Benchmarking and evaluating large-multimodal image generation with lmms.
\newblock 2025.
\newblock URL \url{https://api.semanticscholar.org/CorpusID:277741112}.

\bibitem[Wang et~al.(2023)Wang, Tan, Bi, Xu, Luan, Sunkavalli, Wang, Xu, and Zhang]{wang2023pf}
Peng Wang, Hao Tan, Sai Bi, Yinghao Xu, Fujun Luan, Kalyan Sunkavalli, Wenping Wang, Zexiang Xu, and Kai Zhang.
\newblock Pf-lrm: Pose-free large reconstruction model for joint pose and shape prediction.
\newblock \emph{arXiv preprint arXiv:2311.12024}, 2023.

\bibitem[Wang et~al.(2024{\natexlab{a}})Wang, Bai, Wang, Qin, Chen, Li, Tang, and Hu]{wang2024instantid}
Qixun Wang, Xu~Bai, Haofan Wang, Zekui Qin, Anthony Chen, Huaxia Li, Xu~Tang, and Yao Hu.
\newblock Instantid: Zero-shot identity-preserving generation in seconds.
\newblock \emph{arXiv preprint arXiv:2401.07519}, 2024{\natexlab{a}}.

\bibitem[Wang et~al.(2024{\natexlab{b}})Wang, Fu, Huang, He, and Jiang]{wang2024ms}
Xierui Wang, Siming Fu, Qihan Huang, Wanggui He, and Hao Jiang.
\newblock Ms-diffusion: Multi-subject zero-shot image personalization with layout guidance.
\newblock \emph{arXiv preprint arXiv:2406.07209}, 2024{\natexlab{b}}.

\bibitem[Wang et~al.(2024{\natexlab{c}})Wang, Zhang, Luo, Sun, Cui, Wang, Zhang, Wang, Li, Yu, Zhao, Ao, Min, Li, Wu, Zhao, Zhang, zi~Wang, Liu, He, Yang, Liu, Lin, Huang, and Wang]{Wang2024Emu3NP}
Xinlong Wang, Xiaosong Zhang, Zhengxiong Luo, Quan Sun, Yufeng Cui, Jinsheng Wang, Fan Zhang, Yueze Wang, Zhen Li, Qiying Yu, Yingli Zhao, Yulong Ao, Xuebin Min, Tao Li, Boya Wu, Bo~Zhao, Bowen Zhang, Lian zi~Wang, Guang Liu, Zheqi He, Xi~Yang, Jingjing Liu, Yonghua Lin, Tiejun Huang, and Zhongyuan Wang.
\newblock Emu3: Next-token prediction is all you need.
\newblock \emph{ArXiv}, abs/2409.18869, 2024{\natexlab{c}}.
\newblock URL \url{https://api.semanticscholar.org/CorpusID:272968818}.

\bibitem[Wei et~al.(2025)Wei, Zheng, Zhang, Liu, Ji, Zhang, and Zuo]{wei2025personalized}
Yuxiang Wei, Yiheng Zheng, Yabo Zhang, Ming Liu, Zhilong Ji, Lei Zhang, and Wangmeng Zuo.
\newblock Personalized image generation with deep generative models: A decade survey.
\newblock \emph{arXiv preprint arXiv:2502.13081}, 2025.

\bibitem[Wiles et~al.(2024)Wiles, Zhang, Albuquerque, Kaji{\'c}, Wang, Bugliarello, Onoe, Knutsen, Rashtchian, Pont-Tuset, and Nematzadeh]{wiles2024revisiting}
Olivia Wiles, Chuhan Zhang, Isabela Albuquerque, Ivana Kaji{\'c}, Su~Wang, Emanuele Bugliarello, Yasumasa Onoe, Chris Knutsen, Cyrus Rashtchian, Jordi Pont-Tuset, and Aida Nematzadeh.
\newblock Revisiting text-to-image evaluation with gecko: On metrics, prompts, and human ratings.
\newblock \emph{arXiv preprint arXiv:2404.16820}, 2024.

\bibitem[Xiong et~al.(2024)Xiong, Xiong, Shi, Zhang, Song, and Jacobs]{xiong2024groundingbooth}
Zhexiao Xiong, Wei Xiong, Jing Shi, He~Zhang, Yizhi Song, and Nathan Jacobs.
\newblock Groundingbooth: Grounding text-to-image customization.
\newblock \emph{arXiv preprint arXiv:2409.08520}, 2024.

\bibitem[Yarom et~al.(2023)Yarom, Bitton, Changpinyo, Aharoni, Herzig, Lang, Ofek, and Szpektor]{Yarom2023WhatYS}
Michal Yarom, Yonatan Bitton, Soravit Changpinyo, Roee Aharoni, Jonathan Herzig, Oran Lang, Eran.~O. Ofek, and Idan Szpektor.
\newblock What you see is what you read? improving text-image alignment evaluation.
\newblock \emph{ArXiv}, abs/2305.10400, 2023.
\newblock URL \url{https://api.semanticscholar.org/CorpusID:258740893}.

\bibitem[Ye et~al.(2023)Ye, Zhang, Liu, Han, and Yang]{ye2023ip}
Hu~Ye, Jun Zhang, Sibo Liu, Xiao Han, and Wei Yang.
\newblock Ip-adapter: Text compatible image prompt adapter for text-to-image diffusion models.
\newblock \emph{arXiv preprint arXiv:2308.06721}, 2023.

\bibitem[Yu et~al.(2022)Yu, Xu, Koh, Luong, Baid, Wang, Vasudevan, Ku, Yang, Ayan, Hutchinson, Han, Parekh, Li, Zhang, Baldridge, and Wu]{Yu2022ScalingAM}
Jiahui Yu, Yuanzhong Xu, Jing~Yu Koh, Thang Luong, Gunjan Baid, Zirui Wang, Vijay Vasudevan, Alexander Ku, Yinfei Yang, Burcu~Karagol Ayan, Ben Hutchinson, Wei Han, Zarana Parekh, Xin Li, Han Zhang, Jason Baldridge, and Yonghui Wu.
\newblock Scaling autoregressive models for content-rich text-to-image generation.
\newblock \emph{Trans. Mach. Learn. Res.}, 2022, 2022.
\newblock URL \url{https://api.semanticscholar.org/CorpusID:249926846}.

\bibitem[Zhang et~al.(2023)Zhang, Xu, Barnes, Zhou, Liu, Zhang, Amirghodsi, Lin, Shechtman, and Shi]{Zhang_2023_ICCV}
Lingzhi Zhang, Zhengjie Xu, Connelly Barnes, Yuqian Zhou, Qing Liu, He~Zhang, Sohrab Amirghodsi, Zhe Lin, Eli Shechtman, and Jianbo Shi.
\newblock Perceptual artifacts localization for image synthesis tasks.
\newblock In \emph{Proceedings of the IEEE/CVF International Conference on Computer Vision (ICCV)}, pages 7579--7590, October 2023.

\bibitem[Zhang et~al.(2025)Zhang, Kou, Wang, Li, Sun, Wang, Li, Wang, Cao, Min, Liu, and Zhai]{Zhang2025QEval100KEV}
Zicheng Zhang, Tengchuan Kou, Shushi Wang, Chunyi Li, Wei Sun, Wei Wang, Xiaoyu Li, Zongyu Wang, Xuezhi Cao, Xiongkuo Min, Xiaohong Liu, and Guangtao Zhai.
\newblock Q-eval-100k: Evaluating visual quality and alignment level for text-to-vision content.
\newblock \emph{ArXiv}, abs/2503.02357, 2025.
\newblock URL \url{https://api.semanticscholar.org/CorpusID:276775486}.

\bibitem[Zheng et~al.(2024)Zheng, Teng, Yang, Wang, Chen, Gu, Dong, Ding, and Tang]{Zheng2024CogView3FA}
Wendi Zheng, Jiayan Teng, Zhuoyi Yang, Weihan Wang, Jidong Chen, Xiaotao Gu, Yuxiao Dong, Ming Ding, and Jie Tang.
\newblock Cogview3: Finer and faster text-to-image generation via relay diffusion.
\newblock \emph{ArXiv}, abs/2403.05121, 2024.
\newblock URL \url{https://api.semanticscholar.org/CorpusID:268297194}.

\bibitem[Zong et~al.(2024)Zong, Jiang, Ma, Song, Shao, Shen, Liu, and Li]{zong2024easyref}
Zhuofan Zong, Dongzhi Jiang, Bingqi Ma, Guanglu Song, Hao Shao, Dazhong Shen, Yu~Liu, and Hongsheng Li.
\newblock Easyref: Omni-generalized group image reference for diffusion models via multimodal llm.
\newblock \emph{arXiv preprint arXiv:2412.09618}, 2024.

\end{thebibliography}
\bibliographystyle{plainnat}
\newpage
\appendix
\section{Appendix}
\subsection{Qualitative Results of \NAME Data}
\begin{figure}[htbp]
    \centering
    \includegraphics[width=\linewidth]{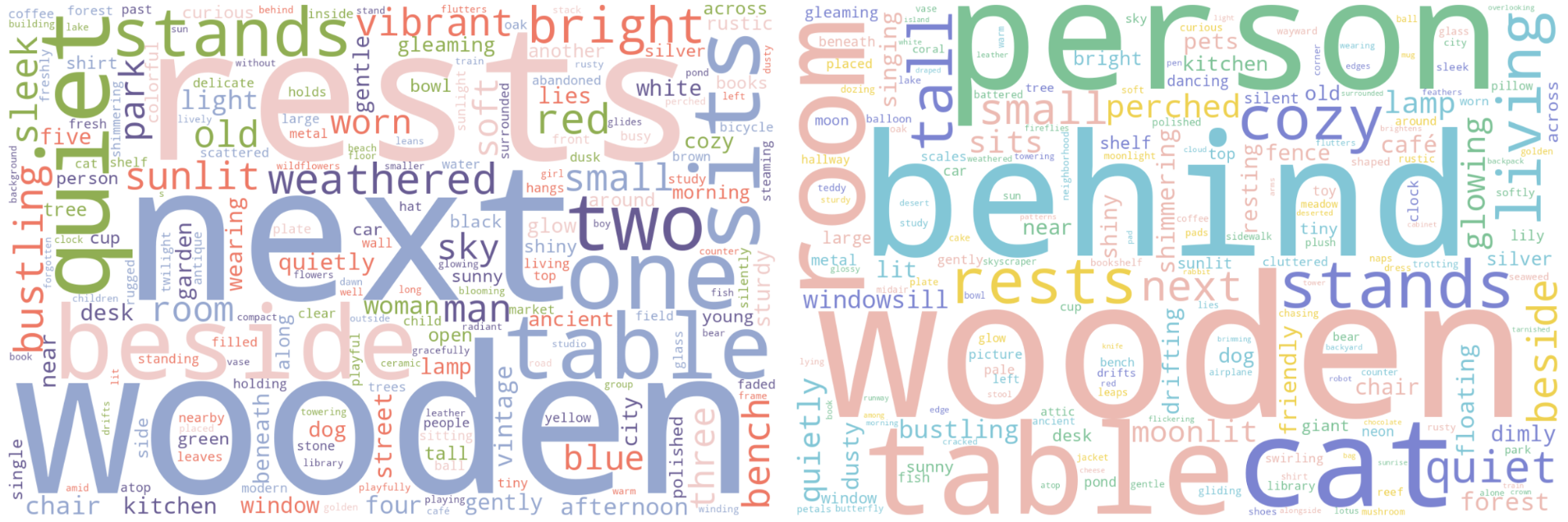}
    \caption{Word clouds of text prompts for the text-only generation (T2I) task (left) and the multimodal generation task (right).}
    \label{fig:wc}
\end{figure}
Figure \ref{fig:wc} visually summarizes the prominent semantic elements in the benchmark prompts for text-only (T2I) and multimodal generation tasks. The differentiation of the word clouds reflects task-specific features of \NAME, emphasizing spatial and descriptive details in T2I tasks, while multimodal tasks more frequently involve social and interactive scenarios.

\subsection{Quantitative and Qualitative Results of AMS}

\begin{figure}[htbp]
    \centering
    \includegraphics[width=\linewidth]{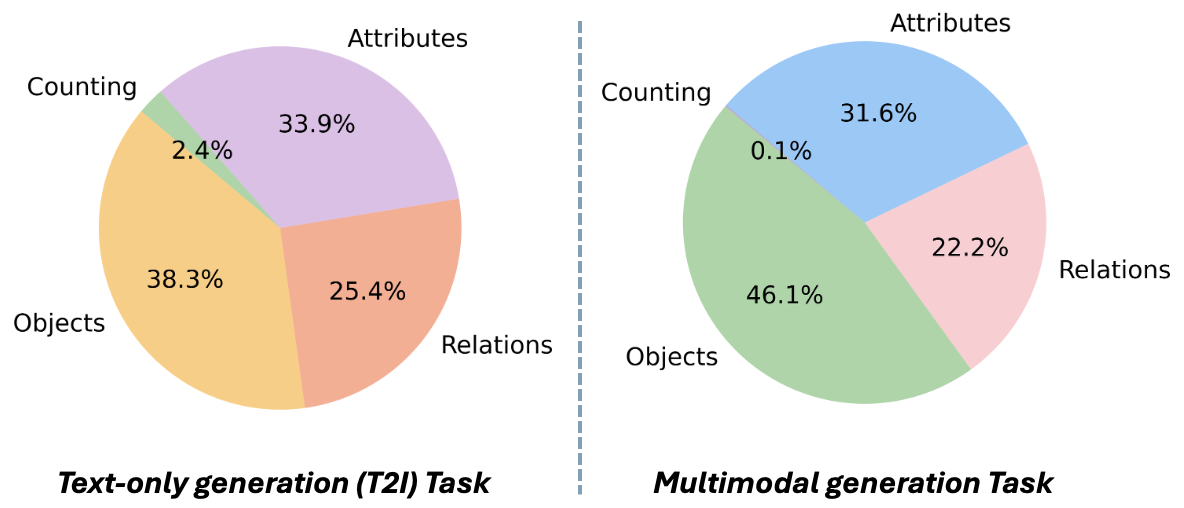}
    \caption{Aspect Distribution of the QA pairs of \textbf{AMS}.}
    \label{fig:aspects}
\end{figure}

\begin{figure}[htbp]
  \centering
  \includegraphics[width=0.49\linewidth]{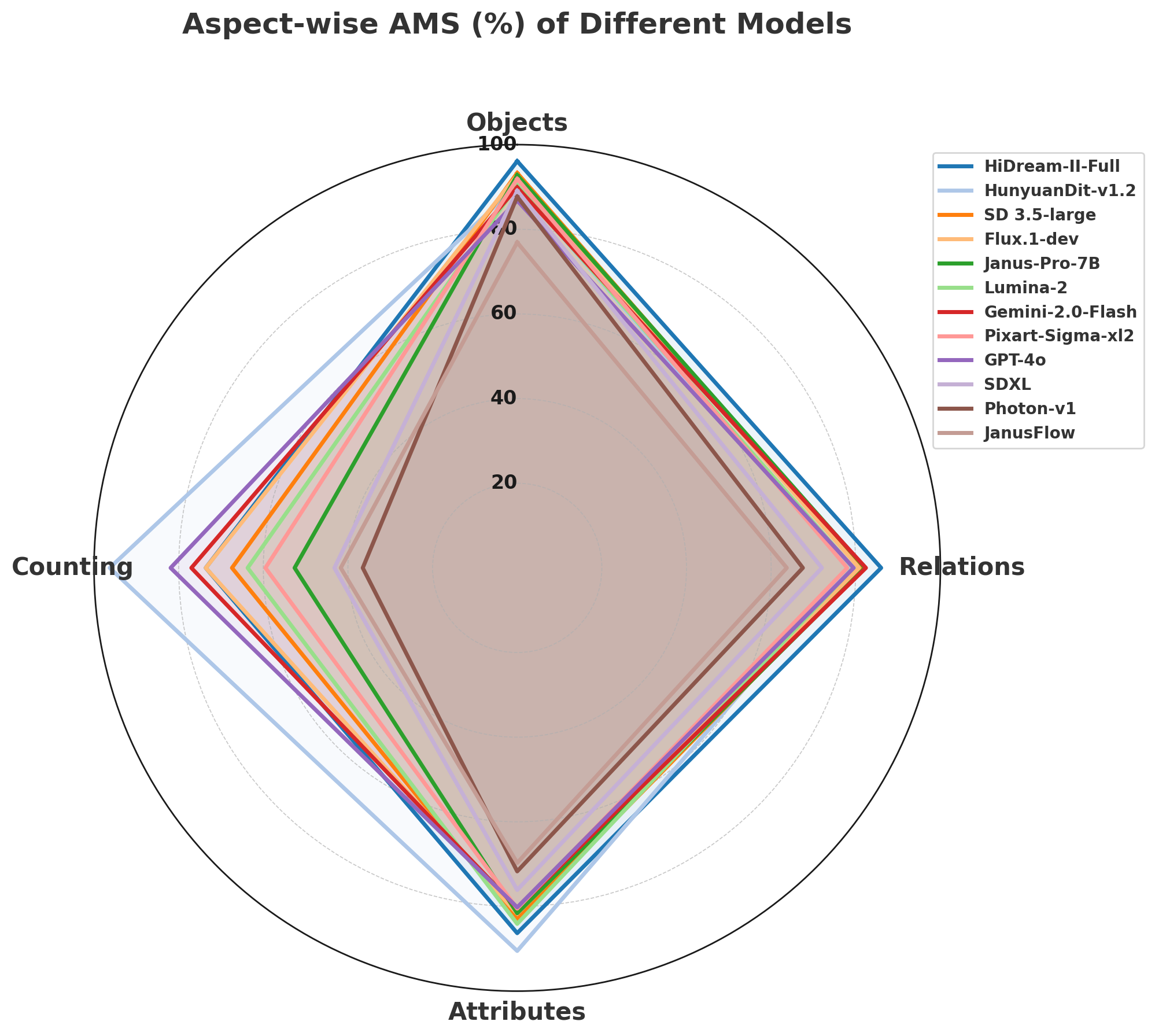}
  \includegraphics[width=0.5\linewidth]{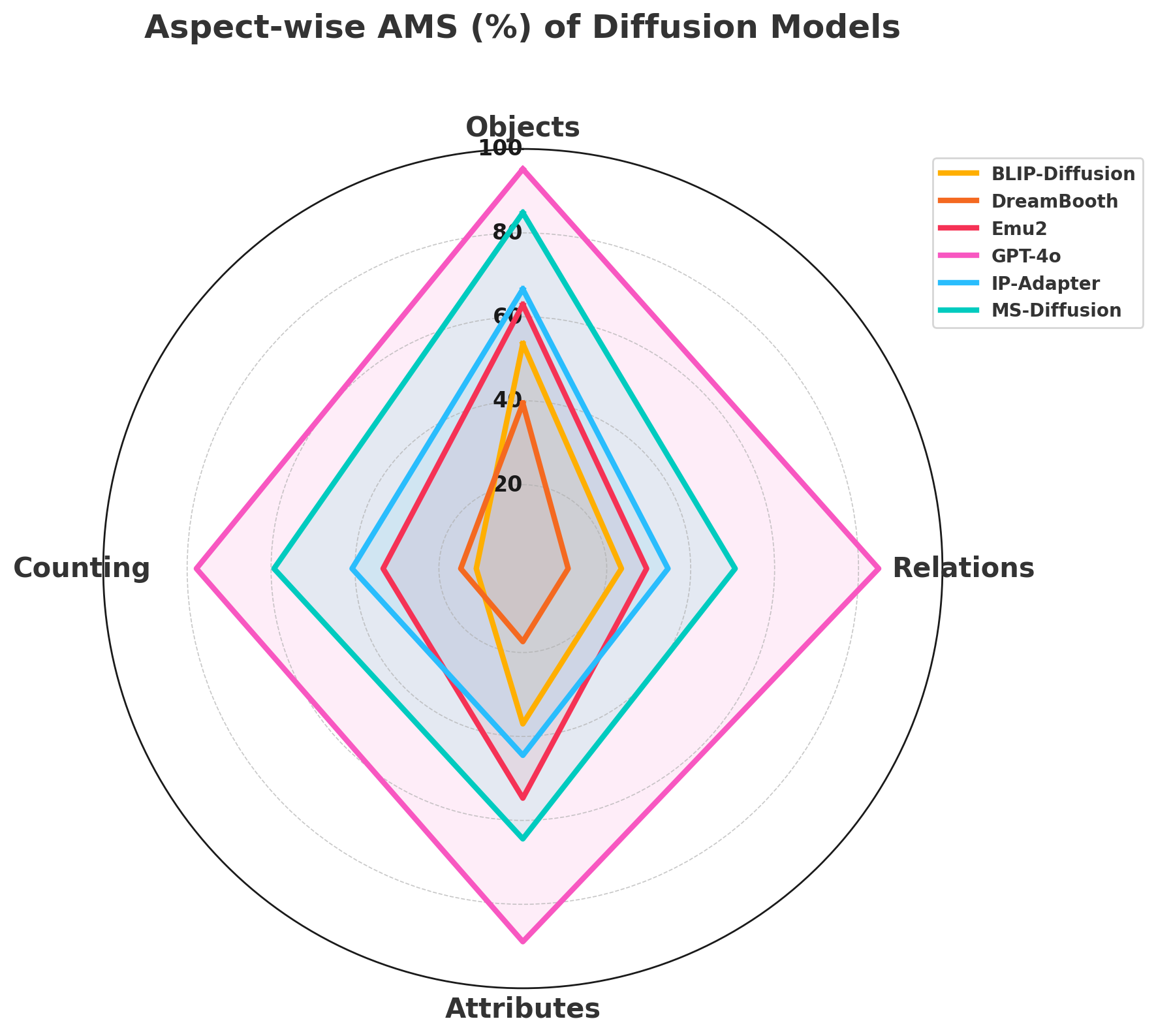}
  \caption{The AMS of different models on the text-only generation (T2I) task (left) and the multimodal generation task (right).}
  \label{fig:radar}
\end{figure}

\begin{table}[htbp]
\centering
\caption{Aspect-level correlation ($\rho$) between \textbf{AMS} and human scores across four aspects.}
\label{tab:aspect_corr}
\resizebox{0.6\linewidth}{!}{%
\begin{tabular}{l|cccc|c}
\toprule
\textbf{Aspect} & \textbf{Objects} $\uparrow$ & \textbf{Relations} $\uparrow$ & \textbf{Attributes} $\uparrow$ & \textbf{Counting} $\uparrow$& \textbf{Overall} $\uparrow$ \\
\midrule
\textbf{Spearman $\rho$} & 0.469 & 0.909 & 0.601 & 0.839&0.699 \\
\bottomrule
\end{tabular}
}

\end{table}

As depicted in Figure \ref{fig:aspects}, the distribution of aspect types differs notably between the text-only generation (T2I) and multi-modal generation tasks. In the T2I setting, “Objects” dominate with 38.3\%, while “Attributes” and “Relations” also constitute substantial proportions (33.9\% and 25.4\%, respectively). In multi-modal generation, “Objects” and “Attributes” remain prominent (46.1\% and 31.6\%, respectively), but the relative proportion of “Relations” decreases significantly (22.2\%). The presence of “Counting” (0.1\%) questions suggests this aspect is less frequent in the customized T2I generation task.

Figure~\ref{fig:radar} presents a comparative analysis of aspect-wise \textbf{AMS} across different models on the text-only generation (T2I) task and the multimodal generation task, highlighting their performance on four key compositional dimensions: Objects, Relations, Attributes, and Counting. On the T2I task, large-scale foundation models such as HiDream-I1, HunyuanDit-v1.2, and SD 3.5-large consistently achieve high AMS scores across aspects, particularly excelling in Objects and Attributes. Specifically, HunyuanDit-v1.2 demonstrates superior Counting performance, underscoring strong numerical understanding in text-driven scenarios. In contrast, for the multimodal generation task, GPT-4o significantly outperforms other diffusion-based models, particularly in complex compositional aspects such as Relations and Counting, highlighting its robust capability in interpreting and synthesizing multimodal inputs. Models like DreamBooth and BLIP-Diffusion show markedly weaker performances, especially in Relations and Counting. These AMS-based comparisons effectively illustrate clear distinctions in compositional understanding capabilities between text-only and multimodal generation settings, emphasizing the metric’s sensitivity in capturing fine-grained model differences.

Table \ref{tab:aspect_corr} further provides quantitative evidence of AMS’s effectiveness: AMS achieves high Spearman correlation with human judgment, particularly in the “Relations” (0.909) and “Counting” (0.839) aspects. This indicates AMS reliably captures complex compositional semantics and aligns closely with human evaluative standards, emphasizing its robustness as a metric for fine-grained image-text alignment evaluation.

\subsection{Experiments Compute Resources}
We conduct our experiments on 8 Nvidia A100 GPUs.

\subsection{Broader Impact}

Multi-modal image generation has wide-ranging applications in areas such as creative design, virtual reality, advertisement, and human-computer interaction. However, the powerful capabilities of these models also pose potential risks, particularly in generating toxic, biased, or harmful visual content.
For instance, the human-centric images in our benchmark could be misused to produce misleading or inappropriate material. \NAME aims to support fair and responsible research by providing a diverse and high-quality dataset while actively mitigating these risks. To this end, we apply thorough filtering to remove toxic, sensitive, or low-quality content from our benchmark. Nevertheless, we encourage the community to consider ethical implications when developing and deploying such models and benchmarks.

\subsection{Instruction Templates for Prompt Generation}

We carefully design eight instruction templates to generate prompts that encompass compositionality, common sense, and diverse stylistic variations. For example, the first template follows a fixed structure: \texttt{[scene description] + [attribute][entity1]
+ [interaction (spatial or action)] + [attribute][entity2]}, which guides GPT-4o to produce prompts that include background context, objects, attributes, and relations. In later templates, we provide GPT-4o with detailed instructions and examples to encourage the generation of prompts that are natural, imaginative, professionally written, or that incorporate elements such as negation, comparison, and numeracy.

\definecolor{codegreen}{rgb}{0,0.6,0}
\definecolor{codegray}{rgb}{0.5,0.5,0.5}
\definecolor{codepurple}{rgb}{0.58,0,0.82}
\definecolor{backcolour}{rgb}{0.95,0.95,0.92}

\lstdefinestyle{mystyle}{
  backgroundcolor=\color{backcolour}, commentstyle=\color{codegreen},
  keywordstyle=\color{magenta},
  numberstyle=\tiny\color{codegray},
  stringstyle=\color{codepurple},
  basicstyle=\ttfamily\footnotesize,
  breakatwhitespace=false,         
  breaklines=true,                 
  captionpos=b,                    
  keepspaces=true,                 
  numbers=left,                    
  numbersep=5pt,                  
  showspaces=false,                
  showstringspaces=false,
  showtabs=false,                  
  tabsize=2
}

\begin{figure}[htbp] 
\centering
\begin{tcolorbox}[colback=white, colframe=SP, text width=0.85\columnwidth, title={\small Instruction Template for T2I Prompts Generation (fixed pattern)}, fontupper=\small, fontlower=\small]
Please generate natural sentences following a format of "[scene description] + [attribute][entity1] + [interaction (spatial or action)] + [attribute][entity2]"; follow the rules below:\\

    1. "entity" should be common objects; e.g., chair, dog, car, lamp, etc. "entity2" is optional. Use "\{entity\}" as entity1 here.\\
    2. "attribute" should be an adjective that describes "shape / color / material / size / condition / etc."\\
    3. "interaction" should describe the relationship between "entity1" and "entity2". "spatial interaction" can be "on the left of / on the right of / on / on top of / on the bottom of / beneath / on the side of / neighboring / next to / touching / in front of / behind / with / etc."; "action interaction" can be any action happening between "entity1" and "entity2", such as "play with, eat, sit, place, hold, etc."\\
    4. "scene description" is the background where the entities appear. It can contain other objects. It is optional.\\
    5. The "interaction action" can be either in active or passive voice.\\
    6. The order of these terms should not be fixed, as long as the sentence still looks natural. E.g., "scene description" can be put at the end.\\
\end{tcolorbox}
\label{template1}
\end{figure}

\begin{figure}[htbp] 
\centering
\begin{tcolorbox}[colback=white, colframe=SP, text width=0.85\columnwidth, title={\small Instruction Template for T2I Prompts Generation (natural)}, fontupper=\small, fontlower=\small]
Please generate prompts in a NATURAL format. It should contain one or more "entities / nouns", (optional) "attributes / adjective" that describes the entities, (optional) "spatial or action interactions" between entities, and (optional) "background description".
Randomly ignore one or more items from [attributes, interactions, background].
One of the entities should be "\{entity\}".
\end{tcolorbox}
\end{figure}

\begin{figure}[htbp] 
\centering
\begin{tcolorbox}[colback=white, colframe=SP, text width=0.85\columnwidth, title={\small Instruction Template for T2I Prompts Generation (unreal)}, fontupper=\small, fontlower=\small]
Please generate prompts in a NATURAL format. It should contain one or more "entities / nouns", (optional) "attributes / adjective" that describes the entities, (optional) "spatial or action interactions" between entities, and (optional) "background description". Note that:\\

    1. Randomly ignore one or more items from [attributes, interactions, background].\\
    2. The description should be imaginative. If imaginative, an example: "A robot and a dolphin dancing under the ocean, surrounded by swirling schools of fish".\\
    3. Avoid repeating sentences you've already generated.
\end{tcolorbox}
\end{figure}

\begin{figure}[htbp] 
\centering
\begin{tcolorbox}[colback=white, colframe=SP, text width=0.85\columnwidth, title={\small Instruction Template for T2I Prompts Generation (professional)}, fontupper=\small, fontlower=\small]
Imagine that you are a professional designer, please write prompt for testing text-to-image diffusion models.
The prompts should look like natural sentences.
Please do not include descriptions about styles, such as "minimalism meets hygge vibes / editorial photoshoot style / baroque detail / etc.".
One of the entities/nouns should be "\{entity\}".
\end{tcolorbox}
\end{figure}

\begin{figure}[htbp] 
\centering
\begin{tcolorbox}[colback=white, colframe=SP, text width=0.85\columnwidth, title={\small Instruction Template for T2I Prompts Generation (negation)}, fontupper=\small, fontlower=\small]
Please generate prompts in a NATURAL format. It should contain one or more "entities / nouns", (optional) "attributes / adjective" that describes the entities, (optional) "spatial or action interactions" between entities, and (optional) "background description". Note that:\\

    1. Randomly ignore one or more items from [attributes, interactions, background].\\
    2. It should include the logic of "negation", such as the examples below:\\
    "The girl with glasses is drawing, and the girl without glasses is singing.",\\
    "In the supermarket, a man with glasses pays a man without glasses.",\\
    "The larger person wears a yellow hat and the smaller person does not.",\\
    "Adjacent houses stand side by side; the left one sports a chimney, while the right one has none.",\\
    "A tailless, not black, cat is sitting.",\\
    "A smiling girl with short hair and no glasses.",\\
    "A bookshelf with no books, only a single red vase.".\\
One of the entities/nouns should be "\{entity\}".
\end{tcolorbox}
\end{figure}

\begin{figure}[htbp] 
\centering
\begin{tcolorbox}[colback=white, colframe=SP, text width=0.85\columnwidth, title={\small Instruction Template for T2I Prompts Generation (comparison)}, fontupper=\small, fontlower=\small]
Please generate prompts in a NATURAL format. It should contain one or more "entities / nouns", (optional) "attributes / adjective" that describes the entities, (optional) "spatial or action interactions" between entities, and (optional) "background description". Note that:\\

    1. Randomly ignore one or more items from [attributes, interactions, background].\\
    2. It should have the logic of "comparison", such as the examples below:\\
    "In a magnificent castle, a red dragon sits and a green dragon flies.",\\
    "A magician holds two books; the left one is open, the right one is closed.",\\
    "One cat is sleeping on the table and the other is playing under the table.".\\
    "A green pumpkin is smiling happily, while a red pumpkin is sitting sadly.",\\
One of the entities/nouns should be "\{entity\}".
\end{tcolorbox}
\end{figure}

\begin{figure}[htbp] 
\centering
\begin{tcolorbox}[colback=white, colframe=SP, text width=0.85\columnwidth, title={\small Instruction Template for T2I Prompts Generation (counting)}, fontupper=\small, fontlower=\small]
Please generate prompts in a NATURAL format. It should contain one or more "entities / nouns", and "numeracy" that describes the number of the entity.\\
Follow the six examples below:\\

    1. four dogs played with two toys.\\
    2. two chickens, four pens and one lemon.\\
    3. Five cylindrical mugs beside two rectangular napkins.\\
    4. three helicopters buzzed over two pillows.\\
    5. Three cookies on a plate.\\
    6. A group of sheep being led by two shepherds across a green field.\\
Avoid repeating sentences you've already generated.
\end{tcolorbox}
\end{figure}

\begin{figure}[htbp] 
\centering
\begin{tcolorbox}[colback=white, colframe=SP, text width=0.85\columnwidth, title={\small Instruction Template for T2I Prompts Generation (numeracy in fixed structure)}, fontupper=\small, fontlower=\small]
Please generate natural sentences following a format of "[scene description (optional)] + [number][attribute][entity1] + [interaction (spatial or action)] + [number (optional)][attribute][entity2]"; follow the rules below:\\

    1. "entity" should be common objects; e.g., chair, dog, car, lamp, etc. "entity2" is optional. Use "{entity}" as entity1 here.\\
    2. "attribute" should be an adjective that describes "shape / color / material / size / condition / etc."\\
    3. "number" should be "two/three/four/..." before the attribute, indicating the number of entities. It is optional for entity2.\\
    4. "interaction" should describes the relationship between "entity1" and "entity2". "spatial interaction" can be "on the left of / on the right of / on / on top of / on the bottom of / beneath / on the side of / neighboring / next to / touching / in front of / behind / with / and / etc."; "action interaction" can be any action happening between "entity1" and "entity2", such as "play with, eat, sit, place, hold, etc."\\
    5. "scene description" is the background where the entities appear. It can contain other objects. It is optional.\\
    6. The "interaction action" can be either in active or passive voice.\\
    7. The order of these terms should not be fixed, as long as the sentence still looks natural. E.g., "scene description" can be put at the end.\\
\end{tcolorbox}
\end{figure}

\begin{figure}[htbp] 
\centering
\begin{tcolorbox}[colback=white, colframe=SP, text width=0.85\columnwidth, title={\small Prompt Template for Text Prompts Aspect Extraction}, fontupper=\small, fontlower=\small]
You need to analyze the query to a aspect graph that matches all the objects, relations (e.g.,spatial relations, action, complex relation), attributes, and counting (number of objects). \\
                    Please ignore all the redundant phrases that are irrelevant to the contents of the image in the query, for example, 'a photo/picture of something, 'something in the background' etc., should not appear in the parsed graph.\\
                    Please also remove all the redundent aspects in the parsed graph.
                    Here are some examples, if there are no such aspect, you can use an empty list to represent:\\
                    For the counting information, please ignore the object numbers that less than 2 (<2).\\
                    
                    Context: \\
                    
                    A group of women is playing the piano in the room.\\
                    \{'Objects':['woman','room'],\\
                    'Other Relations':['play piano'], \\
                    'Spatila Relations':['in, (the room)'], \\
                    'Attributes':[], \\
                    'Counting':['a group of, (Non-specific quantity of woman)']\}\\

                    Two Chihuahuas run after a child on a bicycle.\\
                    \{'Objects':['Chihuahua','child','bicycle'],\\
                    'Other Relations':['runs after, (Chihuahua runs after child)','ride, (ride by the child)'], \\
                    'Spatila Relations':['on, (child on bicycle)']\\
                    'Attributes':['Chihuahua, (Chihuahua is a breed of dog)'], \\
                    'Counting':[Two (number of Chihuahua)]\}
                    \}\\

                    A Delta Boeing 777 taxiing on the runway.\\
                    \{'Objects':['Delta Boeing 777','runway'],\\
                    'Other Relations':['taxiing on, ( the runway)'], \\
                    'Spatial Relations':['on (plane on the runway)'], \\
                    'Attributes':['None'], 'Counting':[]\}\\
                    
                    Please extrace all the aspects precisely!
\end{tcolorbox}
\end{figure}

\begin{figure}[htbp] 
\centering
\begin{tcolorbox}[colback=white, colframe=SP, text width=0.85\columnwidth, title={\small Prompt Template for AMS QA Pair Generation}, fontupper=\small, fontlower=\small]
Given an image and its corresponding caption, generate Visual Question Answering pairs that assess the presence of specific objects, attributes, relations, and counting information in the image.\\
The questions should be phrased naturally, appropriate, and reasonable.
Input:\\
Caption: Two dogs are fighting over a red Frisbee that is bent in half.
Target Elements:\\
\{\{"Objects": ["dog", "Frisbee"], "Relations": ["fighting over, (dogs fighting over Frisbee)"], "Attributes": ["red, (color of Frisbee)", "bent in half, (condition of Frisbee)"], "Counting": ["two, (number of dogs)"]\}\}\\

Example Output (JSON):\\
\{\{"question": "Is there a dog in the image?", "answer": "Yes", "Aspect":'Objects'\}\},\\
    \{\{"question": "Is there a Frisbee in the image?", "answer": "Yes","Aspect":'Objects'\}\},\\
    \{\{"question": "Are the dogs fighting over a Frisbee?", "answer": "Yes","Aspect":'Relations'\}\},\\
    \{\{"question": "Is the Frisbee red?", "answer": "Yes","Aspect":'Attributes'\}\},\\
    \{\{"question": "Is the Frisbee bent in half?", "answer": "Yes","Aspect":'Relations'\}\},\\
    \{\{"question": "Are there two dogs in the image?", "answer": "Yes","Aspect":'Counting'\}\}\\
If the counting aspect is related to 'one, (number of something)', please ignore it!\\
Please reduce the redundancy of the questions, don't repeat !\\
If the question includes relational references—such as friend, mother, daughter, etc.—please specify the associated referent (for example, the woman's friend).\\
If the aspect entity has no practical significance, please ignore it.\\
Input:\\
    Caption: \\
    Target Elements: \\
Output (JSON):
\\
\end{tcolorbox}
\end{figure}

\subsection{Text-Image-Conditioned Dataset Overview}

\begin{figure}[htbp]
    \centering
    \includegraphics[width=\linewidth]{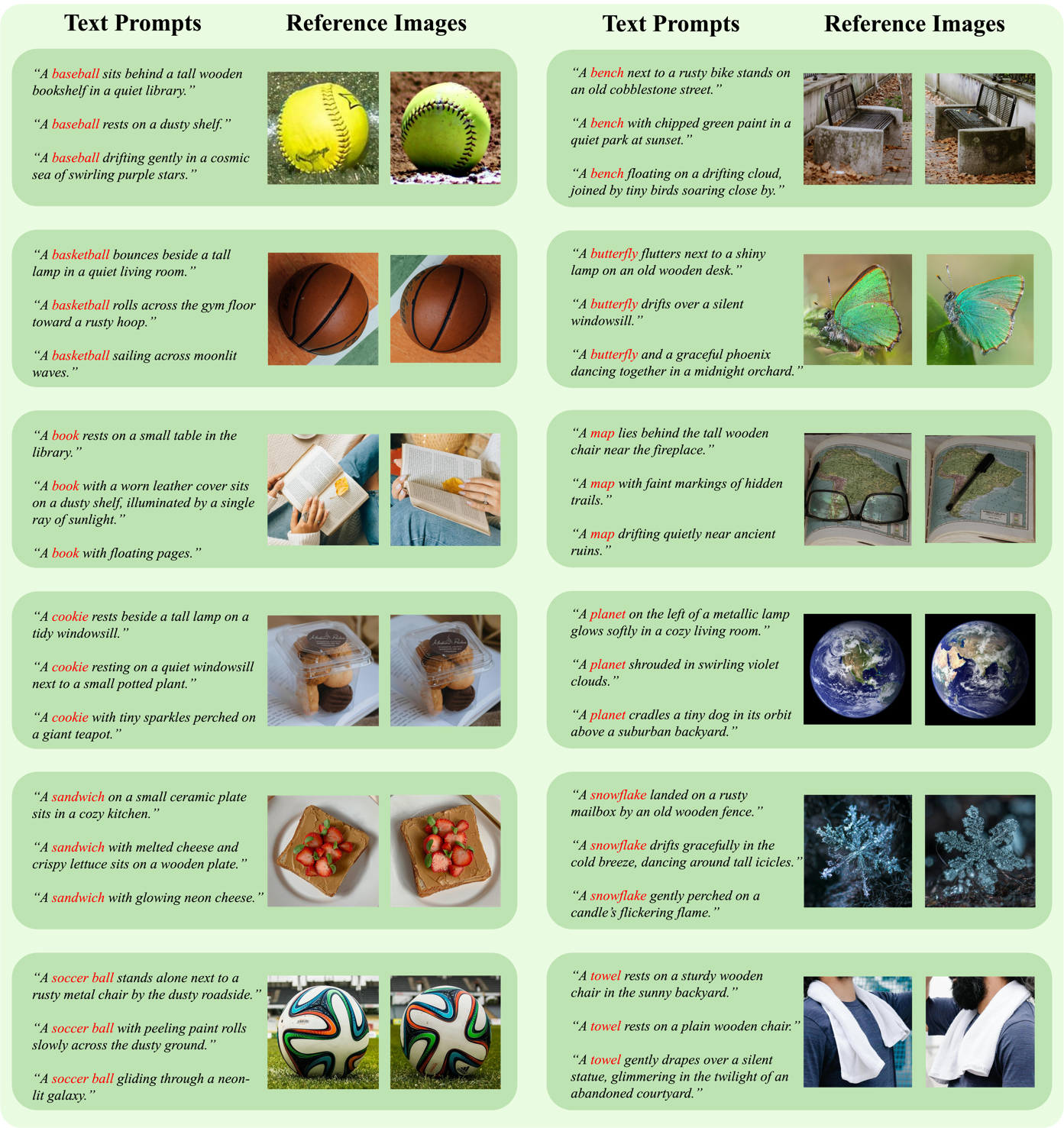}
    \caption{Overview of \NAME.}
    \label{fig:dataset_overview}
\end{figure}

An overview of our comprehensive \NAME is shown in Fig.~\ref{fig:dataset_overview}. Based on the 207 common entities we curated, we collect 386 reference image groups, each containing 3–5 multi-view, object-centric images, and generate 4,850 text prompts that include these entities. The prompts are densely labeled and exhibit rich, detailed semantics, covering compositionality, common sense, and styles.

\subsection{More Qualitative Results}

\begin{figure}[htbp]
    \centering
    \includegraphics[width=\linewidth]{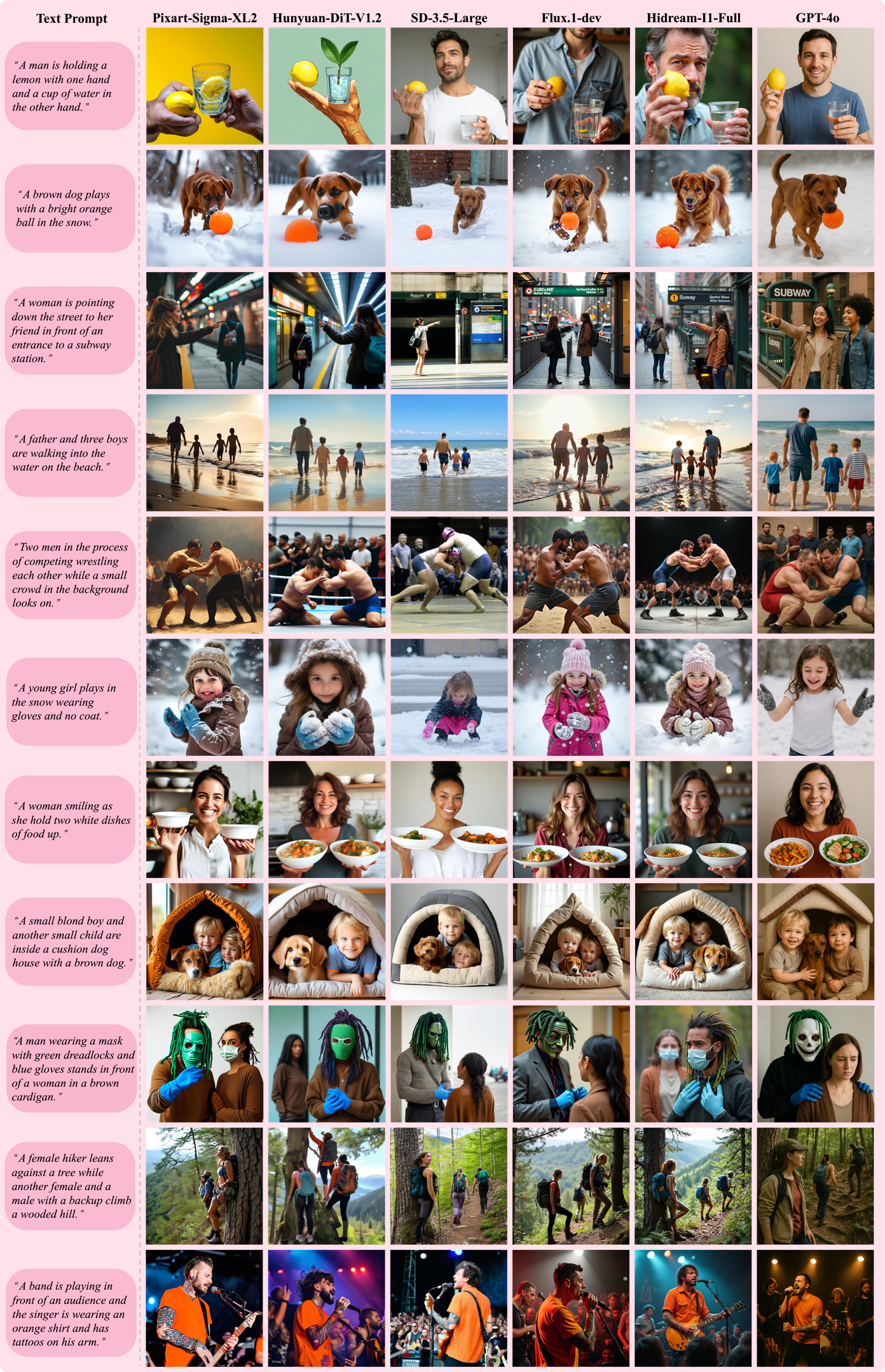}
    \caption{More qualitative results of text-only generation methods on \NAME.}
    \label{fig:qual_supp1}
\end{figure}

\begin{figure}[htbp]
    \centering
    \includegraphics[width=\linewidth]{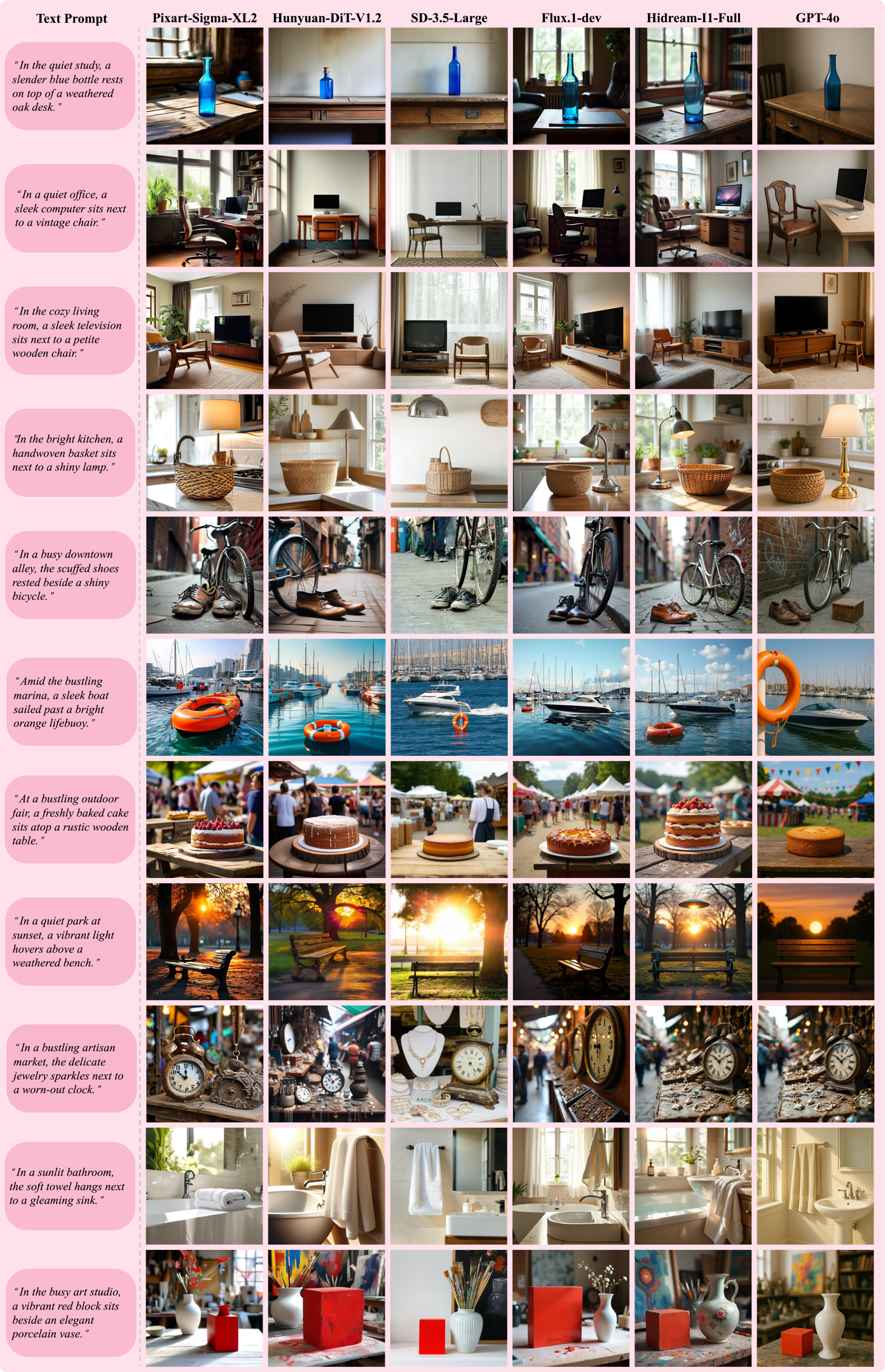}
    \caption{More qualitative results of text-only generation methods on \NAME.}
    \label{fig:qual_supp2}
\end{figure}

\begin{figure}[htbp]
    \centering
    \includegraphics[width=\linewidth]{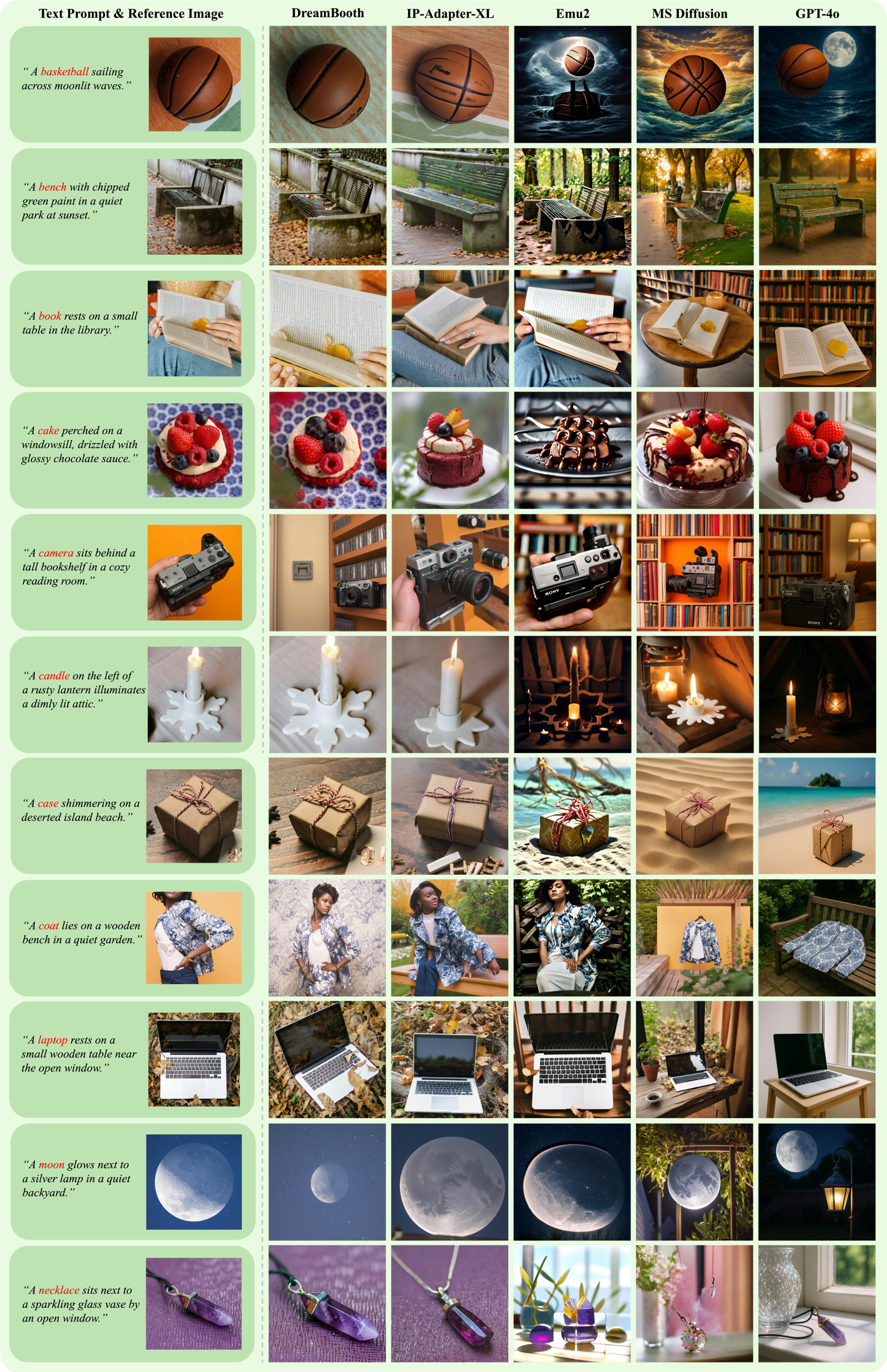}
    \caption{More qualitative results of text-image-conditioned generation methods on \NAME.}
    \label{fig:qual_supp3}
\end{figure}

We show more visual comparisons of the state-of-the-art models in Fig.~\ref{fig:qual_supp1}, \ref{fig:qual_supp2} and \ref{fig:qual_supp3}.

\subsection{Human Evaluation Interface}

\begin{figure}[htbp]
    \centering
    \includegraphics[width=0.85\linewidth]{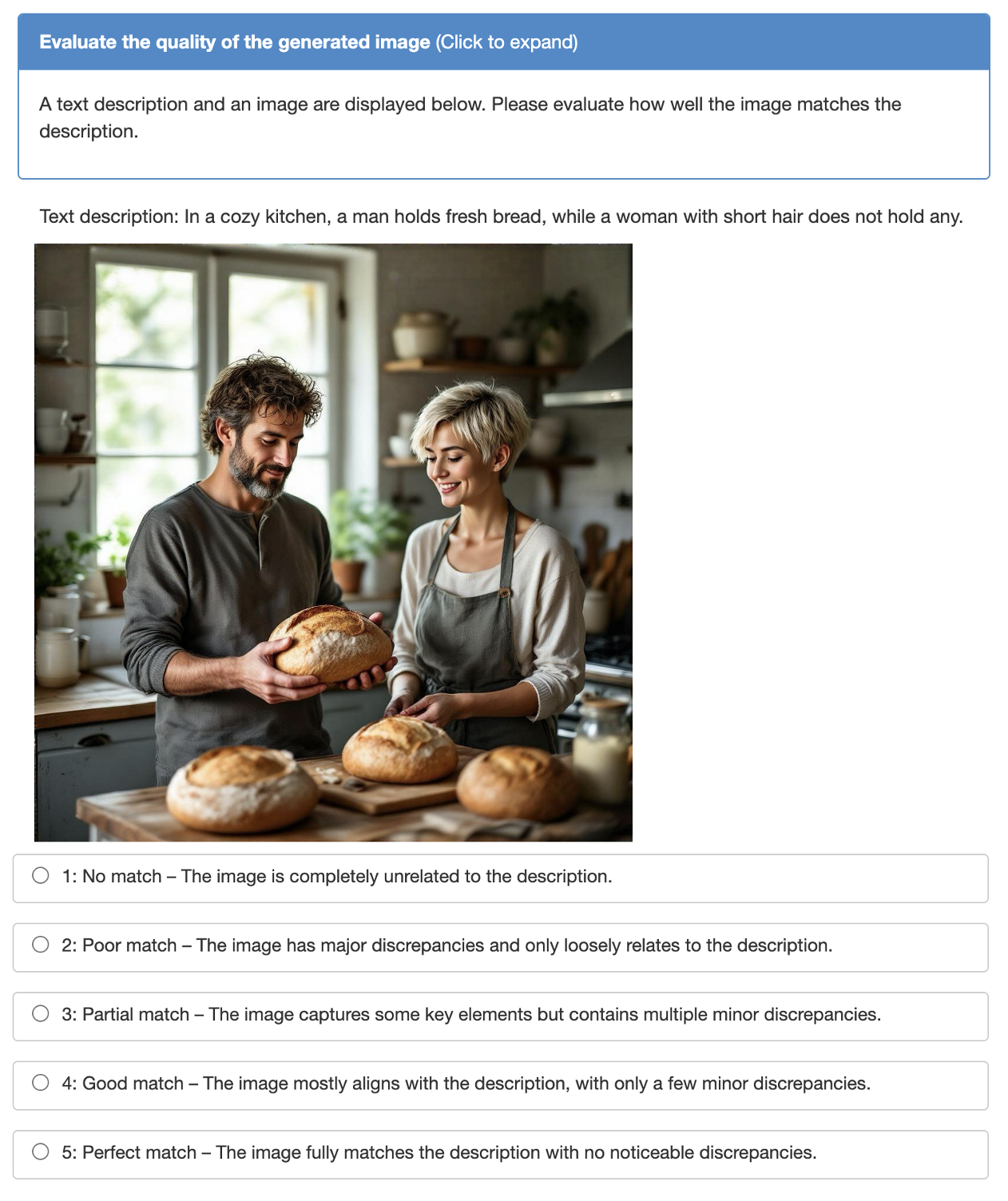}
    \caption{The interface of user study for general prompt following.}
    \label{fig:user_study_1}
\end{figure}

\begin{figure}[htbp]
    \centering
    \includegraphics[width=0.85\linewidth]{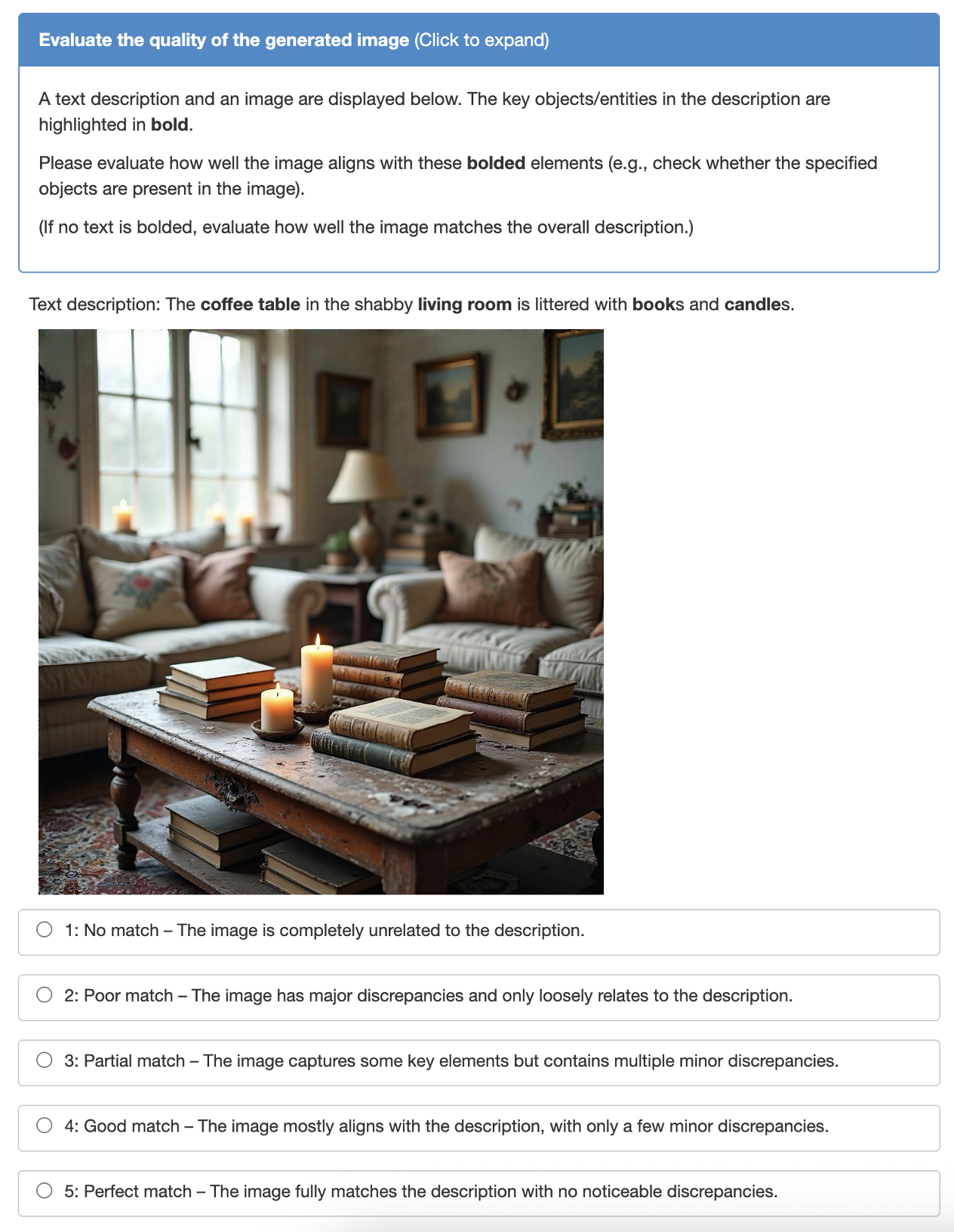}
    \caption{The interface of user study for prompt following on \textit{Object}.}
    \label{fig:user_study_2}
\end{figure}

\begin{figure}[htbp]
    \centering
    \includegraphics[width=0.85\linewidth]{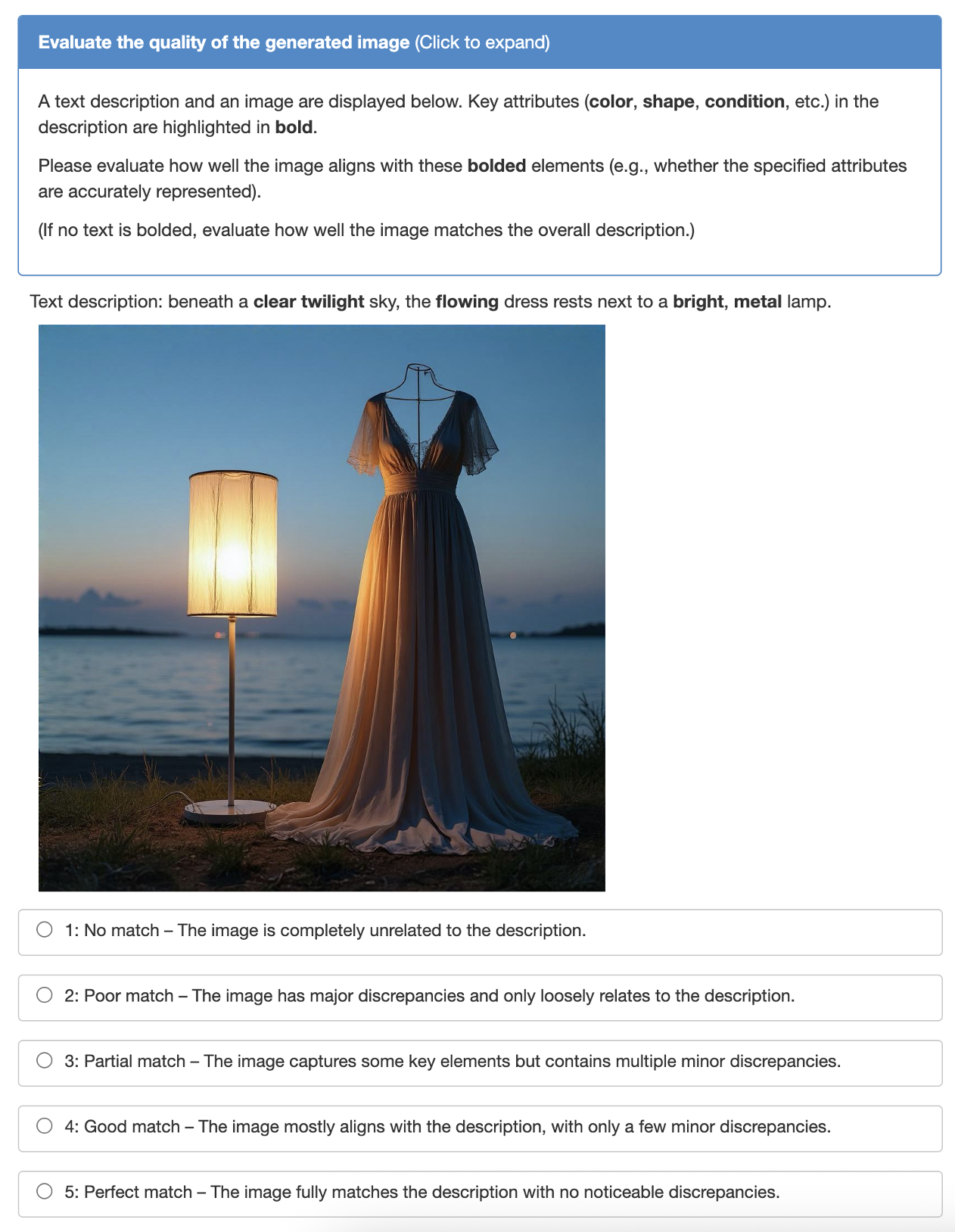}
    \caption{The interface of user study for prompt following on \textit{Attributes}.}
    \label{fig:user_study_3}
\end{figure}

\begin{figure}[htbp]
    \centering
    \includegraphics[width=0.85\linewidth]{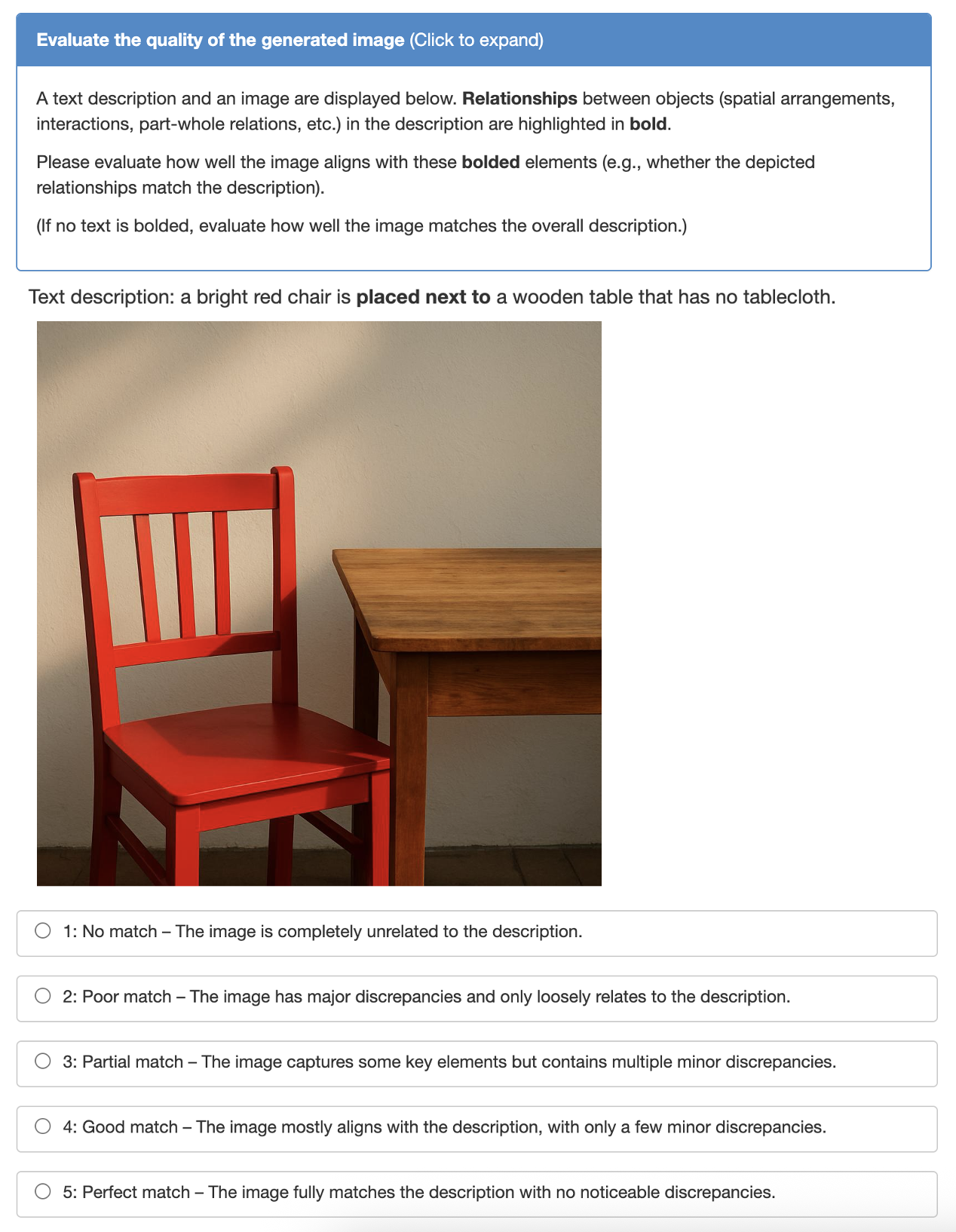}
    \caption{The interface of user study for prompt following on \textit{Relations}.}
    \label{fig:user_study_4}
\end{figure}

\begin{figure}[htbp]
    \centering
    \includegraphics[width=0.85\linewidth]{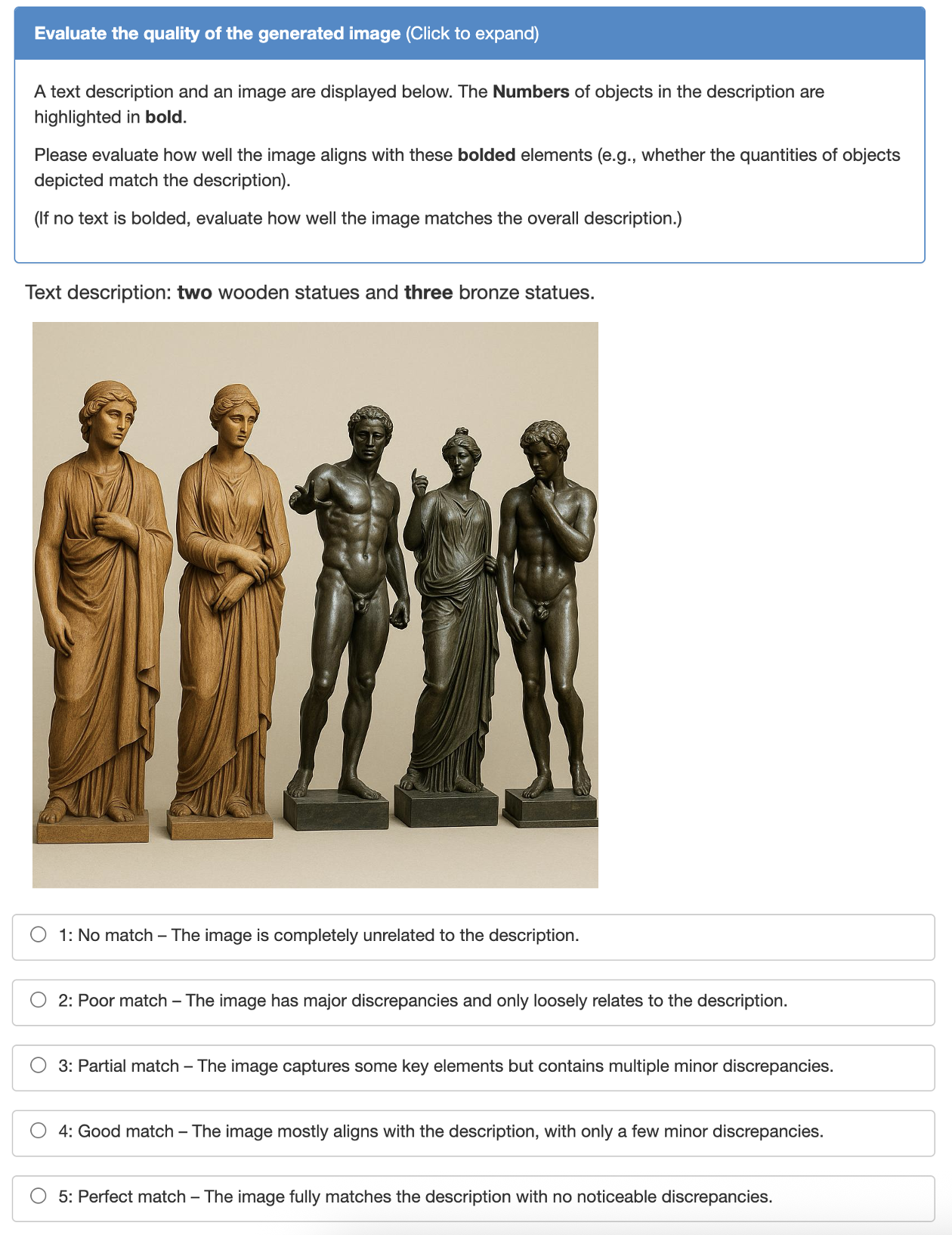}
    \caption{The interface of user study for prompt following on \textit{Numeracy}.}
    \label{fig:user_study_5}
\end{figure}

The Amazon Mechanical Turk interfaces used in the user studies are shown in Fig.~\ref{fig:user_study_1}-\ref{fig:user_study_5}. The study is divided into five categories to assess the compositionality of prompt-image alignment across different aspects: general prompt following (Fig.~\ref{fig:user_study_1}), object (Fig.~\ref{fig:user_study_2}), attribute (Fig.~\ref{fig:user_study_3}), relation (Fig.~\ref{fig:user_study_4}) and numeracy (Fig.~\ref{fig:user_study_5}). In each session, a randomly selected prompt-image pair is presented to the user, who is then asked to rate the generation quality using a 5-point scale. Each question is independently rated by three different workers to ensure reliability.

\end{document}